\pdfoutput=1

\documentclass[prodmode,acmtalip]{acmsmall}

\usepackage{amsmath}
\usepackage{CJK}
\usepackage[T1]{fontenc}
\usepackage{tikz}
\usepackage[CJKbookmarks]{hyperref}

\makeatletter
\AtBeginDvi{\pdfmapline{=tsuku@Unicode@ <tsuku.ttf}}
\DeclareFontFamily{C70}{tsuku}{\hyphenchar \font\m@ne}
\DeclareFontShape{C70}{tsuku}{l}{n}{ <-> CJK * tsuku}{}
\DeclareFontShape{C70}{tsuku}{m}{n}{ <-> CJK * tsuku}{\CJKnormal}
\DeclareFontShape{C70}{tsuku}{bx}{n}{ <-> CJKb * tsuku}{\CJKbold}
\makeatother

\newcommand{\eids}[1]{{\begin{CJK}{UTF8}{tsuku}#1\end{CJK}}}

\usetikzlibrary{arrows,calc,decorations.markings,decorations.pathreplacing,%
  shapes.callouts,shapes.geometric}

\usepgflibrary{arrows.new}

\tikzset{bigah/.style={draw,ultra thick,->,>=latex' new,arrow head=6mm}}

\begin{document}

\markboth{M. Skala}{A Structural Query System for Han Characters}

\title{A Structural Query System for Han Characters}
\author{MATTHEW SKALA
\affil{University of Manitoba}}


\begin{abstract}
The IDSgrep structural query system for Han character dictionaries is
presented.  This system includes a data model and syntax for describing the
spatial structure of Han characters using Extended Ideographic Description
Sequences (EIDSes) based on the Unicode IDS syntax; a language for querying
EIDS databases, designed to suit the needs of font developers and foreign
language learners; a bit vector index inspired by Bloom filters for faster
query operations; a freely available implementation; and format translation
from popular third-party IDS and XML character databases.  Experimental
results are included, with a comparison to other software used for similar
applications.
\end{abstract}


\acmformat{Matthew Skala, 2014.
A Structural Query System for Han Characters.}

\begin{bottomstuff}
Author's address: Matthew Skala, Department of Computer Science,
E2--445 EITC, University of Manitoba, Winnipeg MB R3T 2N2, Canada.
\end{bottomstuff}

\maketitle



\section{Introduction}
\label{sec:intro}

Systematic efforts to describe and categorize Han characters date back as
far as the Second Century~\cite{Creamer:Shuowen}.  Creamer describes
\emph{Shouwen Jiezi}, a very early Chinese dictionary, which divided its
character set into 540 headings, most corresponding to semantic
components (radicals) that might appear in the characters. 
Details like the specific list of radicals have varied over time and with
the differing purposes for which dictionaries have been compiled, but that
general scheme remained the standard for hardcopy dictionaries of Han
characters until the Twentieth Century.  A user looking up an unknown
character would start by identifying the radical (a task
made easier by native language knowledge, and experience with the specific
dictionary's classification scheme) and then search the appropriate section,
which might be further organized by number of strokes.

More recent hardcopy dictionaries, especially those aimed at foreign
language learners, have used other organizational schemes.  For instance,
the SKIP method~\cite{Halpern:NJECD} uses an easily-memorized numerical
description defined by the visual appearance, not the semantics, of the
character.  A user can identify the dictionary head for \eids{}\eids{明} as
``1--4--4,'' meaning ``divided into left and right parts, with four strokes
on the left and four on the right,'' and find the character in the
dictionary without needing to know which side means ``moon.''

Computerized dictionaries offer many search methods.  WWWJDIC~\cite{WWWJDIC}
is one of the best-known.  As a Web-based resource it is constantly updated,
and it offers queries by traditional radical-stroke classification; by SKIP
code; by several other dictionary classification schemes; and by
cross-reference to other dictionaries and character lists (including Unicode
code points).  Its interactive searches are especially convenient for users
who do not know the Han character set well.  In multi-radical mode, the user
chooses one component at a time, from a list that is roughly the few hundred
radicals of traditional dictionaries, but notably without any requirement
to choose the single official radical of each character.  A
dynamically-updated list shows all characters in the database that contain
all (by simple Boolean AND) of the chosen components.  In handwriting
recognition mode, the user can write the first few strokes of a character
using the mouse, with a dynamic list updating to show the closest matches in
the database to the strokes written so far.  Usually, the desired character
will move to the top of the list after a few strokes; but the user must know
enough about Han script to be able to guess the stroke order.  These kinds
of interactive searches represent the current state of the art in
widely-deployed computerized character dictionaries.

In this paper we describe the IDSgrep structural query system for Han
character databases.  IDSgrep originated as an internal development tool for
the Tsukurimashou Project's Japanese-language parametric font
family~\cite{Skala:Tsukurimashou}.  Tsukurimashou represents characters as
subroutines written in a fully featured programming language, with the
structure of the code echoing the visual appearance, not the semantic
organization, of the characters.  For instance, the code for \eids{明}
invokes subroutines for \eids{日} and \eids{月} and a subroutine that
abstracts the operation of placing components in a left-right configuration. 
As a matter of software design, when writing new character definitions the
developer must be able to find other characters (both those already in the
fonts and those that may be added in the future) with similar structures
that could share code.

Queries with specific geometric constraints like ``which other characters,
if any, have the same right-hand side as this one?'' are not easy to answer
with traditional dictionaries that focus more on meaning than on spatial
organization.  Negative search results are of interest, and when the font
project hopes to cover obscure characters not known to most native readers,
negative results from a native reader's domain knowledge are only of limited
use.  A human expert can at best say ``I can't think of a character fitting
this description,'' but to be sure that \emph{there exists no} character
fitting a given description we need a precise semantics of character
descriptions, a database we can trust to contain all characters of interest,
and a tool for querying the database.  IDSgrep is the query tool; it defines
at least a syntax for the descriptions; and it can make use of existing
databases that are complete enough to be useful.
With a contrasting approach from existing dictionaries, IDSgrep may find
application outside the original scope of font development.

The present paper's contributions are a data model and query language for
the spatial structure of Han characters adapted to dictionary use;
algorithmic techniques for efficient implementation of the query language;
and the experimental evaluation of a practical implementation.  The software
is freely available from the Tsukurimashou Project's Web site on Sourceforge
Japan~\cite{Tsukurimashou}.

\subsection{Character description languages}

Computer typesetting projects for Han-script languages have long used
descriptions of the character glyphs in terms of smaller components, with
varying degrees of complexity and formal specification in how those
components may be combined.  Some work in this area has focused on
Knuth's~\citeyear{Knuth:METAFONT}
METAFONT system, in which glyphs to be typeset are
described using a fully powered computer programming language and components
and combining operations can be invoked as subroutines.  Many authors have
worked on METAFONT-related Han script projects over the course of more
than three decades, with the Tsukurimashou Project that gave birth to
IDSgrep as one of the most recent
contributions~\cite{Mei:LCCD,Hobby:Chinese,Hosek:Design,Yiu:Chinese,Laguna:Hong,Skala:Tsukurimashou}. 
The Wadalab font project~\cite{Tanaka:Wadalab} implemented similar concepts
using LISP instead of METAFONT, and was one of the most successful projects
of its kind; fonts it generated are in wide use in the free software
community to this day.  Any such project implicitly extends the programming
language used into a language for describing Han characters, but most do not
treat the descriptions as separate entities from the software code. 
HanGlyph~\cite{Yiu:Chinese} is one exception: it defines a formal syntax for
a description language that is translated by separate and
character-independent software.

Several projects use XML rather than a programming language to describe
characters, and these projects often emphasize dictionary and database
applications instead of primarily font creation.  Font creation may
nonetheless be included as one intended application of the data.  Such
projects include Structural Character Modeling Language
(SCML)~\cite{Peebles:SCML}, Character Description Language
(CDL)~\cite{Wu:CDL}, GlyphWiki~\cite{Kamichi:GlyphWiki}, and
KanjiVG~\cite{KanjiVG}.  Here the focus is often on providing high-quality
data in a convenient form for application development, with such details as
user interface and query language left to the application developers to
determine.  Although IDSgrep does not query XML directly, it is one such
query application.  The possibility of using the popular XML databases, and
KanjiVG in particular, was one factor motivating its design.

\subsection{Tree searching}

The general problem of searching for a pattern in a large input is one of
the most thoroughly studied in computer science.  Searching utilities like
GNU grep~\cite{grep} are widely used.  At least among expert users,
grep-like regular expression search is regarded as the standard for flexible
text searching and is expected as a standard feature of text editors,
database software, and programming languages or libraries.  The desire to
apply something similar to Han characters motivated IDSgrep development:
``why not run grep on the writing system itself?''~\cite{Skala:Tsukurimashou}

Considered as a general-purpose searching utility, IDSgrep does something
much like regular expression matching on tree structures.  Regular
expression matching generalized to trees, and other kinds of tree pattern
matching, have been studied both as abstract
problems~\cite{Aiken:Implementing} and with specific application to
searching parse trees in computational linguistics
applications~\cite{Lai:Querying}.  The Tregex utility~\cite{Levy:Tregex} is
a popular implementation in the computational linguistics domain, used for
comparison in the experimental section of the present work.

Although the system can process other kinds of queries too, many important
IDSgrep queries take the form of an example tree with some parts left as
match-anything wildcards.  The matching operation on such a query is
equivalent to the \emph{unification} operation on terms in logic programming
languages like Prolog~\cite{Clocksin:Prolog}, and algorithmic techniques
applicable to unification are of interest for IDSgrep and IDSgrep-like tree
matching.

Unification can also be defined in a lattice of types, and
one well-known technique for unification in type lattices
represents the types as bit vectors with bitwise AND and zero-testing to
represent the unification operation~\cite{AitKaci:Efficient}.  The bit
vector approach to type unification has been extended to generalize the zero
value, which permits the use of shorter vectors and thus faster
processing~\cite{Skala:Generalized}; and to permit approximate results via
the Bloom filter concept~\cite{Bloom:Space,Skala:Approximate}, allowing
further speed improvement when the bit vector test is used as a guard for a
more expensive non-approximate test.  The present work applies similar ideas
to speed improvement for tree matching.  The work of \citeN{Kaneta:Faster}
on unordered pseudo-tree matching with bit vectors is also of interest; it
considers a very different tree-matching problem, but it uses some similar
bit-vector techniques, and has a strong theoretical analysis.


\section{The EIDS data model and syntax}
\label{sec:data-model}

Unicode defines a simple grammar for describing Han characters
as strings called \emph{Ideographic Description Sequences}
(IDSes)~\cite{Unicode:IDS}.  An IDS
is one of the following:
\begin{itemize}
  \item a single character chosen from a set that includes the
    Unicode-encoded Han characters, strokes for building up Han characters,
    and radicals or components that may occur in Han characters;
  \item one of the prefix binary operators \eids{⿰⿱⿴⿵⿶⿷⿸⿹⿺⿻}
    followed by two more IDSes, defined recursively; or
  \item one of the prefix ternary operators \eids{⿲⿳} followed by three
    IDSes, defined recursively.
\end{itemize}

\begin{figure}
  \begin{tikzpicture}[baseline=(current bounding box.north)]
    \node at (0,1.3) {\scalebox{2}{\eids{明→⿰日月}}};
    \node at (0,0) {\scalebox{3}{\eids{⿰}}};
    \node at (-0.8,-1.6) {\scalebox{3}{\eids{日}}};
    \node at (0.8,-1.6) {\scalebox{3}{\eids{月}}};
    \draw[>=latex',->,ultra thick] (-0.25,-0.5) -- (-0.55,-1.1);
    \draw[>=latex',->,ultra thick] (0.25,-0.5) -- (0.55,-1.1);
  \end{tikzpicture}
  \hspace{\fill}
  \begin{tikzpicture}[baseline=(current bounding box.north)]
    \node at (0,1.3) {\scalebox{2}{\eids{語→⿰言⿱五口}}};
    \node at (0,0) {\scalebox{3}{\eids{⿰}}};
    \node at (-0.8,-1.6) {\scalebox{3}{\eids{言}}};
    \node at (0.8,-1.6) {\scalebox{3}{\eids{⿱}}};
    \draw[>=latex',->,ultra thick] (-0.25,-0.5) -- (-0.55,-1.1);
    \draw[>=latex',->,ultra thick] (0.25,-0.5) -- (0.55,-1.1);
    \node at (0,-3.2) {\scalebox{3}{\eids{五}}};
    \node at (1.6,-3.2) {\scalebox{3}{\eids{口}}};
    \draw[>=latex',->,ultra thick] (0.55,-2.1) -- (0.25,-2.7);
    \draw[>=latex',->,ultra thick] (1.05,-2.1) -- (1.35,-2.7);
  \end{tikzpicture}
  \hspace{\fill}
  \begin{tikzpicture}[baseline=(current bounding box.north)]
    \node at (0,1.3) {\scalebox{2}{\eids{⿴囗⿱⿰木山丁}}};
    \node at (0,0) {\scalebox{3}{\eids{⿴}}};
    \node at (-0.8,-1.6) {\scalebox{3}{\eids{囗}}};
    \node at (0.8,-1.6) {\scalebox{3}{\eids{⿱}}};
    \draw[>=latex',->,ultra thick] (-0.25,-0.5) -- (-0.55,-1.1);
    \draw[>=latex',->,ultra thick] (0.25,-0.5) -- (0.55,-1.1);
    \node at (0,-3.2) {\scalebox{3}{\eids{⿰}}};
    \node at (1.6,-3.2) {\scalebox{3}{\eids{丁}}};
    \draw[>=latex',->,ultra thick] (0.55,-2.1) -- (0.25,-2.7);
    \draw[>=latex',->,ultra thick] (1.05,-2.1) -- (1.35,-2.7);
    \node at (-0.8,-4.8) {\scalebox{3}{\eids{木}}};
    \node at (0.8,-4.8) {\scalebox{3}{\eids{山}}};
    \draw[>=latex',->,ultra thick] (-0.25,-3.7) -- (-0.55,-4.3);
    \draw[>=latex',->,ultra thick] (0.25,-3.7) -- (0.55,-4.3);
  \end{tikzpicture}
  \caption{Sample Unicode IDSes and their associated trees.}
  \label{fig:unicode-eids}
\end{figure}
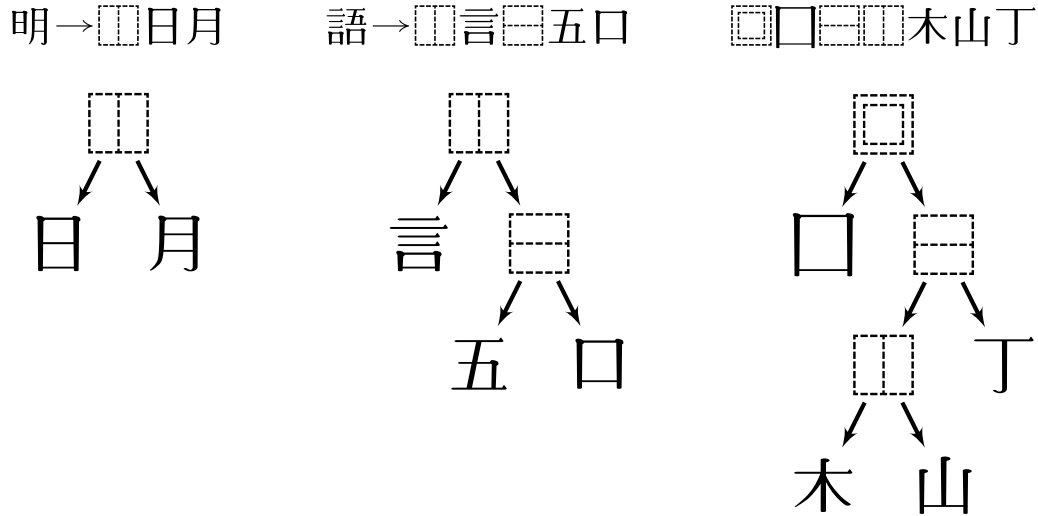

Example Unicode IDSes include ``\eids{⿰日月}'' for ``\eids{明}'';
``\eids{⿰言⿱五口}'' for ``\eids{語}''; and
``\eids{⿴囗⿱⿰木山丁}'' for an unencoded nonsense character.  These are
shown in Figure~\ref{fig:unicode-eids}.

The binary and ternary operators (Unicode uses the term ``trinary'') are
special characters defined for this purpose, with code points in the range
U+2FF0 to U+2FFB.  Their exact semantics are not precisely defined by
Unicode, but are at least suggested by the associated names and graphical
symbols.  It is understood that in ordinary situations they should be
displayed as graphical characters; they are not \emph{combining} or
\emph{control} characters in the Unicode-related technical senses of those
terms.  Earlier versions of Unicode imposed limits on the maximum total
length of an IDS and the number of consecutive non-operator characters
permitted to occur in an IDS, but both are unlimited in the current version. 
The limits were intended to make it easier for software to find the start
and end of an IDS without looking too far forward or backward in a stream of
characters.

Many aspects of when and how to use IDSes are understood to be
application-dependent and left unspecified.  The standard non-bindingly
encourages the use of IDSes that are as short as possible, which implies
simply using the encoded character for any character that has an encoding,
and using encoded characters to represent the largest possible components of
unencoded characters rather than breaking them down further.  For some
applications, such as dictionaries, it may nonetheless be desirable to break
down encoded characters to a finer level.  That question of how finely to
break down character components motivates one of the significant extensions
introduced by the IDSgrep system, namely the ``head'' concept.

IDSgrep describes characters using \emph{Extended Ideographic Description
Sequences} (EIDSes), which are strings of Unicode characters expressing
abstract data structures called \emph{EIDS trees}~\cite{IDSgrep:Manual}.  We
define the EIDS trees first.  An EIDS tree is a tree data structure with the
following properties.
\begin{itemize}
  \item Each node has a \emph{functor}, which is a nonempty
    string of Unicode characters.
  \item Each node may optionally have a \emph{head}, which if present is a
    nonempty string of Unicode characters.
  \item Each node has a sequence of between zero and three children,
    which are EIDS trees defined recursively.
\end{itemize}

The number of children of a node is called its \emph{arity}.  Functors, and
heads where present, usually consist of single characters, but that is not a
requirement.

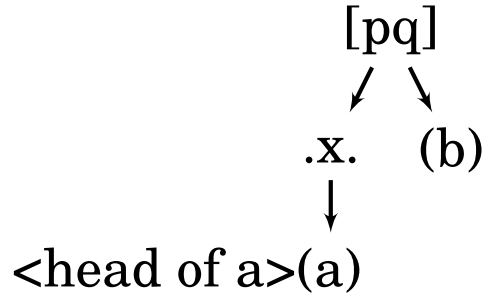
\begin{figure}
  \centering
  \begin{tikzpicture}
    \node at (-1.5,1.3) {\scalebox{2}{\eids{[pq].x.<head of a>(a)(b)}}};
    \node at (0,0) {\scalebox{2}{\eids{[pq]}}};
    \node at (-0.8,-1.6) {\scalebox{2}{\eids{.x.}}};
    \node at (0.8,-1.6) {\scalebox{2}{\eids{(b)}}};
    \draw[>=latex',->,ultra thick] (-0.25,-0.5) -- (-0.55,-1.1);
    \draw[>=latex',->,ultra thick] (0.25,-0.5) -- (0.55,-1.1);
    \node[anchor=east] at (-0.27,-3.2) {\scalebox{2}{\eids{<head of a>(a)}}};
    \draw[>=latex',->,ultra thick] (-0.8,-2.0) -- (-0.8,-2.7);
  \end{tikzpicture}
  \caption{A sample EIDS and its EIDS tree.}
  \label{fig:arity-example}
\end{figure}

The most explicit EIDS character string for a given EIDS tree consists of
the head of the root enclosed in ASCII angle brackets \eids{<>}, or omitted
if the root has no head; the functor of the root, enclosed in parentheses
\eids{()}, dots \eids{..}, square brackets \eids{[]}, or curly
braces \eids{\{\}} for arity zero, one, two, or three respectively; and
then the EIDSes for all the root's children, recursively.  For instance, an
EIDS in this explicit syntax might be written ``\eids{[pq].x.<head of
a>(a)(b)}''.  The associated EIDS tree is shown in
Figure~\ref{fig:arity-example}.

However, the syntax includes several additional features designed both to
make it easier to use and to allow valid Unicode IDSes to be valid IDSgrep
EIDSes.  The fully bracketed form would rarely be used in practice.  First,
all the Unicode IDS operator characters such as \eids{⿱} and
\eids{⿰}, and some special characters used in IDSgrep pattern matching,
are considered to have implicit brackets of the appropriate type when they
occur where an opening bracket would otherwise appear.  These are called
\emph{sugary implicit brackets} (from the term ``syntactic
sugar''~\cite{Landin:Mechanical}).  For
instance, ``\eids{⿰(a)(b)}'' expresses the same EIDS tree as
``\eids{[⿰](a)(b)}''.

If a character is not an opening bracket itself, is not on the short list of
characters with sugary implicit brackets, and does not have some other
special function (such as backslash for character-code escapes), then by
default the character is considered to have implicit \eids{<>} head brackets
and also be followed by ``\eids{(;)}'', a \emph{syrupy implicit semicolon}. 
Han characters and their components fall into this category.  Thus a single
character like ``\eids{語}'' is a valid EIDS as well as a valid Unicode IDS,
and parsing it produces the same EIDS tree as the explicitly bracketed
``\eids{<語>(;)}''.

A few other syntax rules exist, covering issues like backslash escapes;
ASCII aliases of the Unicode IDS operators to allow them to be more easily
typed on an ASCII keyboard; and non-ASCII aliases for the bracket characters
to allow more visually appealing formatting of dictionary entries.  These
points are beyond the scope of the current discussion, but described in the
IDSgrep documentation~\cite{IDSgrep:Manual}.  One remaining rule significant
to the current work is that because neither a head nor a functor may be
empty, if the closing bracket that would end a bracketed string occurs
immediately after the opening bracket---which would otherwise create an
illegal empty string---then it does \emph{not} end the string but becomes
the first character of the string.  The important, and motivating,
consequence of this rule is that ``\eids{...}'' is valid syntax for the
functor of a unary node consisting of a single ASCII period; that is the
frequently-used ``match anywhere'' operator in the query language of the
next section.

\begin{figure}
  \centering
  \begin{tikzpicture}[yscale=1.3,xscale=2]
    \node at (0.3,1.0) {\scalebox{2}{\eids{<語>⿰言<吾>⿱五口}}};
    \node at (0,0) {\scalebox{2}{\eids{<語>[⿰]}}};
    \node at (-0.8,-1.6) {\scalebox{2}{\eids{<言>(;)}}};
    \node at (0.8,-1.6) {\scalebox{2}{\eids{<吾>[⿱]}}};
    \draw[>=latex',->,ultra thick] (-0.25,-0.4) -- (-0.55,-1.2);
    \draw[>=latex',->,ultra thick] (0.25,-0.4) -- (0.55,-1.2);
    \node at (0,-3.2) {\scalebox{2}{\eids{<五>(;)}}};
    \node at (1.6,-3.2) {\scalebox{2}{\eids{<口>(;)}}};
    \draw[>=latex',->,ultra thick] (0.55,-2.0) -- (0.25,-2.8);
    \draw[>=latex',->,ultra thick] (1.05,-2.0) -- (1.35,-2.8);
  \end{tikzpicture}
  \caption{The EIDS tree for the dictionary entry
    \protect\eids{<語>⿰言<吾>⿱五口}. 
    Note the implicit brackets and semicolons.}
  \label{fig:eids-tree}
\end{figure}
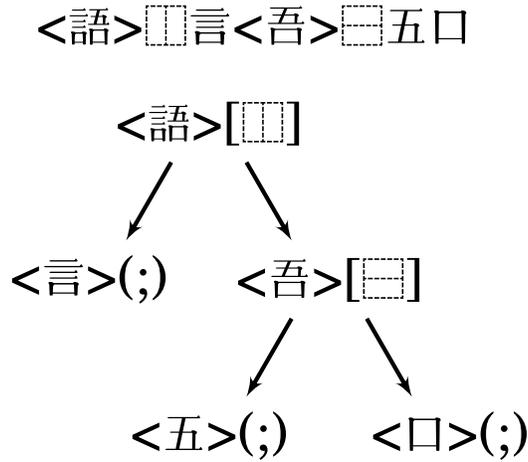

A Unicode IDS maps naturally to the EIDS tree formed by parsing it as an
EIDS.  The IDS operators like \eids{⿸} and \eids{⿳} become the functors of
binary and ternary nodes in the tree under the sugary-bracket rule.  The Han
characters, strokes, and components become the heads of leaf nodes, with
semicolons as their functors, under the syrupy-semicolon rule.  However, it
is also possible to insert heads at other levels of the tree just by
inserting each head in ASCII angle brackets at the appropriate point in the
Unicode IDS.  For instance, a dictionary entry for the character \eids{語}
might look like ``\eids{<語>⿰言<吾>⿱五口}'': the internal
nodes are marked with the characters that represent the subtrees at those
locations even though they are also broken down further.  If a subtree
happened not to be an encoded character in itself, it could be left
anonymous with no head.  A search for this dictionary entry could use the
complete low-level decomposition, or match a subtree or the entire entry by
matching the appropriate head.  The EIDS tree is shown in
Figure~\ref{fig:eids-tree}.

\section{The IDSgrep query language}
\label{sec:query-language}

Just as the Unix grep utility tests each line in its input
against a regular expression and passes the matching lines through to its
output, the IDSgrep command-line utility tests each EIDS in its input
against a matching pattern and passes through those that match.  Because of
the context-free nature of EIDS syntax, traditional regular expressions
would not be sufficient to handle typical queries; instead, IDSgrep defines
its own language of matching patterns in terms of EIDS trees.  Users specify
the matching pattern for a run of IDSgrep by entering it as a string on the
command line in the same syntax used for the dictionary entries.  This
section describes the matching patterns.


\subsection{Query language definition}

Let $\mathcal{E}$ be the set of EIDS trees.  We will define a function
$\mathit{match}:\mathcal{E}\times \mathcal{E}\rightarrow\{\mathsf{T},
\mathsf{F}\}$.  The basic operation of the IDSgrep utility is to parse its
input, evaluate
$\mathit{match}(N,H)$ for one matching pattern or \emph{needle} $N$ specified
on the command line and every matching subject or \emph{haystack} $H$
present in the input, and write to the output every $H$ for which
$\mathit{match}(N,H)=\mathsf{T}$.  The $\mathit{match}$ function is defined
as follows.
\begin{itemize}
  \item If $N$ and $H$ both have heads, then
    $\mathit{match}(N,H)=\mathsf{T}$ if and only if those heads are
    identical.  No other rules are applied.
  \item If $N$ and $H$ do not both have heads, but the functor and arity of
    $N$ are in the set of matching operators $\{ \textrm{\eids{(?)}},
    \textrm{\eids{...}}, \textrm{\eids{.*.}}, \textrm{\eids{.!.}},
    \textrm{\eids{[\&]}}, \textrm{\eids{[|]}}, \textrm{\eids{.=.}},
    \textrm{\eids{.@.}}, \textrm{\eids{./.}}, \textrm{\eids{.\#.}} \}$
    (arities indicated by the brackets around the operators; note \eids{...}
    is one of the operators, not an indication of omitted items), then
    $\mathit{match}(N,H)$ is determined by rules specific to the operator,
    as described below.
  \item Otherwise, $\mathit{match}(N,H)=\mathsf{T}$ if and only if
    $N$ and $H$ have identical functors and arities and
    $\mathit{match}(N_i,H_i)=\mathsf{T}$ recursively for each pair
    $(N_i,H_i)$ of corresponding children of $N$ and $H$.
\end{itemize}

The nullary question mark \eids{(?)} is a match-everything wildcard:
$\mathit{match}(\textrm{\eids{(?)}},H)=\mathsf{T}$ for all $H$.  Three dots
(syntax for a unary functor containing a single dot) match anywhere:
$\mathit{match}(\textrm{\eids{...}}N,H)=\mathsf{T}$ if some subtree of $H$
(possibly all of $H$) is matched by $N$.  The asterisk allows reordering of
children at the top level: $\mathit{match}(\textrm{\eids{.*.}}N,H)=
\mathsf{T}$ if and only if there is some permutation of the children of $N$
that would match $H$.

The basic Boolean operations of NOT, AND, and OR are available through
\eids{.!.}, \eids{[\&]}, and \eids{[|]} respectively.  We have
$\mathit{match}(\textrm{\eids{.!.}}N,H)=\mathsf{T}$ if and only if
$\mathit{match}(N,H)=\mathsf{F}$;
$\mathit{match}(\textrm{\eids{[\&]}}MN,H)=\mathsf{T}$ if and only if
$\mathit{match}(M,H)=\mathsf{T}$ and $\mathit{match}(N,H)=\mathsf{T}$; and
$\mathit{match}(\textrm{\eids{[|]}}MN,H)=\mathsf{T}$ if and only if
$\mathit{match}(M,H)=\mathsf{T}$ or $\mathit{match}(N,H)=\mathsf{T}$.

The equals sign performs literal matching of functors that would otherwise
be interpreted as special.  If $N$ and $H$ both have heads, then
$\mathit{match}(\textrm{\eids{.=.}}N,H)=\mathsf{T}$ if and only if the heads
are identical.  Otherwise,
$\mathit{match}(\textrm{\eids{.=.}}N,H)=\mathsf{T}$ if and only if $N$ and
$H$ have identical functors, identical arities, and
$\mathit{match}(N_i,H_i)=\mathsf{T}$ is true for their corresponding
children.  Those are the same rules as for basic matching without
\eids{.=.}, except that any special matching semantics of the functor of $N$
are ignored.  Trees requiring this operator are not expected to occur in
typical Han character databases, where the functors will normally all be
semicolons and Unicode IDS operators that have no special matching
semantics, but the literal match operator is included because of the basic
design goal that IDSgrep is to be a generic EIDS matching and searching
utility.

The at-sign does rearranged matching for operations governed by an
associative law.  Consider the case of three character components side by
side: it might be written \eids{⿰A⿰BC}, \eids{⿰⿰ABC}, or \eids{⿲ABC}. 
Kawabata proposes normalizing all IDSes into a canonical form (which in this
case would be \eids{⿰A⿰BC}) to make matching
easier~\cite{Kawabata:Normalization}.  But in some applications, such as
describing the structure of code in the Tsukurimashou Project, there may be
meaningful differences between these tree structures, such that it might
sometimes be desired to write a query that matches one of these and not the
others.  It is also a goal of IDSgrep to place as few restrictions as
possible on the form of input trees.  IDSgrep thus implements a special
operator for associative matching, which can be used or not depending on the
desired behaviour of a particular query.  Evaluation of
$\mathit{match}(\textrm{\eids{.@.}}N,H)$ proceeds by starting from the roots
of $N$ and $H$ and descending recursively through all children whose
functors and arities match those of the root.  The remaining subtrees below
the matching nodes are treated as children of notional nodes with unlimited
arity; and then those nodes are compared literally as with the \eids{.=.}
operator (functor and arity must match, and all corresponding children). 
Thus \eids{@⿰⿰AB⿰CD} will match all five cases of \eids{A}, \eids{B},
\eids{C}, and \eids{D} combined in that order by three \eids{⿰} nodes. 
This matching operator does not convert ternary to binary IDS operators, as
the normalization approach would; users must handle that manually if
desired, either using Boolean OR or by suitable normalization of the input
dictionaries.  Note that there is also no special handling of a combination
of \eids{.*.} with \eids{.@.}.

The remaining two special matching operators provide escape from IDSgrep to
other pattern matching systems.  Slash invokes the PCRE regular expression
library~\cite{PCRE}: if $N$ and $H$ both have heads, then
$\mathit{match}(\textrm{\eids{./.}}N,H)=\mathsf{T}$ if and only if the head
of $N$ considered as a PCRE regular expression matches inside the head of
$H$; and if they do not both have heads,
$\mathit{match}(\textrm{\eids{./.}}N,H)=\mathsf{T}$ if and only if $N$ and
$H$ have the same arity, the functor of $N$ considered as a PCRE regular
expression matches inside the functor of $H$, and all children match
recursively.  When dictionary entries come directly from Unicode IDSes and
thus have single-character heads and functors, this operation is unlikely to
be needed; but IDSgrep also has experimental dictionaries that include
definition, pronunciation, and other data in multi-character strings, and
then regular expression searching on those strings can be valuable. 
Finally, the hash operator is for invoking user-defined matching predicates. 
In IDSgrep version 0.5.1 the user-defined predicates test characters against
the coverage of font files; they are not described further here.


\subsection{Examples}

It is expected that in typical use, the IDSgrep utility's main input will be
a dictionary containing the decompositions of characters, with each tree
having a head at the root level containing the character being decomposed,
and then some decomposition below that.  For instance, the IDSgrep
dictionary derived from KanjiVG includes the entry
\eids{<結>⿰糸<吉>⿱士口}.  Note that the subtree
\eids{<吉>⿱士口} has a head of its own, because \eids{吉} is an encoded
character.

The simplest kind of query would then be a single character like \eids{結}. 
Under the parsing rules, that is translated to a tree consisting of a single
nullary (leaf) node with \eids{結} as its head and semicolon as its functor. 
Since all the dictionary entries have heads, matching proceeds by simply
comparing heads for identity; the search will return
\eids{<結>⿰糸<吉>⿱士口} and any other entries that have heads identical to
\eids{結}.  Used this way, IDSgrep performs a simple lookup function.

A more complicated query might specify the complete structure of the
character.  For instance, \eids{⿰糸⿱士口} (note no heads on the non-leaf
nodes) will match \eids{<結>⿰糸<吉>⿱士口} by recursive matching of
subtrees.  In this way IDSgrep might serve to augment an input method:  a
user might know the pronunciation or other information needed to type
\eids{糸}, \eids{士}, and \eids{口}, without knowing how to type \eids{結}. 
Note that the Unicode IDS characters \eids{⿰} and \eids{⿱} can also be
substituted by ASCII aliases \eids{[lr]} and \eids{[tb]}, so need not be
specially supported by the user's input method.

But the main benefit of IDSgrep is for cases where the query only specifies
partial information.  The query \eids{...士} matches all characters that
contain \eids{士} anywhere, with 70 hits in the KanjiVG database.  The query
\eids{\&...士...口} matches all characters containing both \eids{士} and
\eids{口}, with 25 hits in KanjiVG.  These searches mimic the multi-radical
search of many computerized character dictionaries.  IDSgrep can go a step
further than others by capturing spatial information in the query:
\eids{⿰?...士} matches characters that contain \eids{士} as or within the
right side (not just anywhere; 31 hits in KanjiVG), and \eids{⿰?⿱士口}
matches characters that contain \eids{⿱士口} as the right side (6 hits in
KanjiVG).  That latter query might come from a language learner who is
unsure about \eids{糸} but recognizes and can type the other components in
\eids{結}.  A handwriting recognition query would be difficult here because
\eids{糸} comes first in the stroke order, requiring the user to write it
correctly before starting to specify the known components.


\section{Performance enhancements}
\label{sec:enhancements}

A straightforward implementation of the IDSgrep command-line utility might
parse every EIDS tree in its input and then match the trees against the
matching pattern by recursive descent.  That approach is sufficient in the
original application: with the simple queries that tend to occur in
practice, and databases the size of typical dictionaries, the command-line
utility running on a desktop PC can answer queries at about the same speed
that one user can type them.  However, other applications (for instance,
online dictionary servers, linguistics research, and smart-phone dictionary
``apps'') may involve a higher rate of queries or less powerful hardware. 
Such applications motivate further enhancement of IDSgrep's matching
performance.


\subsection{Match filtering}

Recall that EIDS matching defines a function
$\mathit{match}:\mathcal{E}\times \mathcal{E}\rightarrow\{\mathsf{T},
\mathsf{F}\}$ on the set $\mathcal{E}$ of EIDS trees, and the IDSgrep
command-line utility's main purpose is to compute $\mathit{match}(N,H)$ for
a search pattern $N$ and each dictionary entry $H$.  This function takes a
relatively long time to compute.  However, we expect relatively few entries
to actually be matches: with tens of thousands of entries in the dictionary,
a typical query will usually only return a few tens of matches.  It seems
wasteful to run the expensive calculation of $\mathit{match}$ on every
dictionary entry when they will almost all fail, and all the more so because
much of the input (namely, the list of $H$ values) is known in advance and is
the same for every search.  We can reduce the waste and make use of the
advance knowledge of the input by means of \emph{filtering}.

The aim is to reduce the number of times we do the work of
calculating $\mathit{match}(N,H)$, by first calculating some other function
that is much cheaper and will rule out most values of $H$.  If for the large
majority of dictionary entries we can quickly prove that the match will
fail, then we only need calculate $\mathit{match}$ on the few that remain. 
We hope that the time saved by the avoided calls to $\mathit{match}$ will
more than compensate for whatever additional work may be required to
recognize the ruled-out entries.  This general approach is the foundation of
the well-known \emph{Bloom filtering} technique~\cite{Bloom:Space}.

Let $\mathcal{F}$ and $\mathcal{V}$ be sets called the \emph{filters} and
the \emph{vectors} respectively, and define functions
$\mathit{filt}:\mathcal{E}\rightarrow\mathcal{F}$,
$\mathit{vec}:\mathcal{E}\rightarrow\mathcal{V}$, and
$\mathit{check}:\mathcal{F}\times\mathcal{V}\rightarrow\{\mathsf{T},
\mathsf{F}\}$ such that for all $N,H \in \mathcal{E}$, this property holds:
\begin{equation}
  \mathit{match}(N,H)=\mathsf{T} \Rightarrow
    \mathit{check}(\mathit{filt}(N),\mathit{vec}(H))=\mathsf{T} \, .
  \label{eqn:filter-defn}
\end{equation}

We precompute and store the values of $\mathit{vec}(H)$ for each $H$ in the
dictionary.  This is a precomputation done just once for the useful lifespan
of the dictionary, not repeated per query.  To answer a user query $N$, we
compute $\mathit{filt}(N)$, once per query; then
$\mathit{check}(\mathit{filt}(N),\mathit{vec}(H))$, for every query and
every entry.  When
$\mathit{check}(\mathit{filt}(N),\mathit{vec}(H))=\mathsf{F}$, we can skip
to the next entry.  The property \eqref{eqn:filter-defn} guarantees that in
such a case, we know $\mathit{match}(N,H)=\mathsf{F}$ without calculating it
explicitly.  Only when
$\mathit{check}(\mathit{filt}(N),\mathit{vec}(H))=\mathsf{T}$ do we invoke a
more complicated algorithm to compute
$\mathit{match}(N,H)$, and return $H$ as a match to $N$ should that return
$\mathsf{T}$.  The computation of $\mathit{check}$ is time-critical because
it happens for every dictionary entry and every query; increasing the cost
of the other functions, which are invoked fewer times, can be of benefit if
by doing so we can decrease the cost of $\mathit{check}$.

If \eqref{eqn:filter-defn} holds, then the algorithm is correct in the sense
of returning the same set of match results that we would get without
filtering.  Such filtering schemes clearly exist; having $\mathit{check}$
return $\mathsf{T}$ unconditionally is a trivial example.  However, for
filtering to be of benefit, the following properties (paraphrased from the
IDSgrep user manual~\cite{IDSgrep:Manual}) are desirable.  Note that unlike
\eqref{eqn:filter-defn}, which must be absolutely true, it is acceptable for
these properties to hold only on average in common cases.  When they fail,
the system becomes less efficient but remains correct.

\begin{itemize}
  \item Although $\mathit{vec}$ may be expensive to compute, the elements of
    $\mathcal{V}$ it produces as output are small enough that we can afford
    to store them for all dictionary entries.
  \item Although $\mathit{filt}$ may be expensive to compute in comparison
    to $\mathit{check}$, it is still fast enough that we can reasonably
    afford to compute it once for each user-initiated query (each value of
    $N$).
  \item The $\mathit{check}$ function is very fast.
  \item The converse of \eqref{eqn:filter-defn} is usually true, on the
    distribution of search patterns and dictionary entries we expect to see
    in practice.
\end{itemize}

IDSgrep uses two layers of match filtering.  In addition to the dictionary
database containing values of $H$, it reads an index file containing
precomputed values of $\mathit{vec}(H)$.  The two filtering layers share the
definition of $\mathit{vec}$ but differ in their definitions of
$\mathit{filt}$ and $\mathit{check}$.  If the first layer returns
$\mathit{check}(\mathit{filt}(N),\mathit{vec}(H))=\mathsf{F}$ (meaning $H$
is not a potential match), then the command-line utility immediately skips
to the next entry, without checking the second filter.  Only if both filters
return potential matches does the utility read the full dictionary entry
from the main input file, parse the EIDS syntax into a tree data structure,
and compute the full matching function.  The index file has fixed-length
records that include offsets and lengths for the corresponding entries
in the main input, to allow saving on the I/O cost of reading the
variable-length and sometimes large dictionary entries as well as saving on
the parsing and matching operations as such.


\subsection{Bit vectors and $\lambda$ filters}

Classical Bloom filtering is a match filtering scheme much as described
here, applied to subset membership tests.  It is desired to quickly test
whether objects may be elements of some set that was fixed in advance,
without the cost of storing and searching the entire set.  The Bloom filter
applies a small constant number of hash functions to an input object and
uses them as indices into an array of bits.  Each bit is set to $1$ if and
only if any of the hashes applied to any of the objects in the set would
produce that hash value.  When testing an unknown object, we check all its
corresponding bits and return it as a possible match if and only if they are
all $1$.  The scheme may produce some false positives (possible matches that
were not in the set) but no false negatives.  Any object that is in
the set will necessarily return the ``possible match'' result; for any other
object, because we are checking multiple bits that are effectively chosen at
random, as long as the array is large enough to contain a significant
fraction of zero bits, it is reasonably likely but not guaranteed that at
least one of the bits checked will be zero and we can return ``definitely no
match.''  The desired properties hold of recognizing all
objects in the set, and not too many others.  \citeN{Bloom:Space} gives a
detailed analysis, which has become well-known.

It is also well-known that some algebraic operations can be applied to Bloom
filters with useful results: for instance, the bitwise AND of two Bloom
filter bit arrays is a Bloom filter that recognizes the intersection of the
sets they recognize.  \citeN{Guo:Dynamic} give a good summary of results on
algebraic combinations of standard Bloom filters, in the context of
introducing an enhanced version of their own.  IDSgrep uses this algebraic
view of Bloom filters to create a \emph{filter calculus} (called a calculus
because it operates on objects that implicitly represent functions) in
which the filter approximating a complicated EIDS-match query is calculated
from filters that approximate matching its subtrees.  The notion of
\emph{generalized zero}~\cite{Skala:Generalized} detected by counting bits
and testing against a threshold is also applied.

IDSgrep's filter calculus was designed for a specific practical
implementation, and we describe it here as it is implemented in the
software, including the specific parameter values found in IDSgrep version
0.5.1~\cite{IDSgrep:Manual}.  Such things as the vector length could certainly
be changed in other applications, but an attempt to generalize the scheme
and present it without implementation details would not be more easily
understandable.

Let $\mathcal{V}$, the set of vectors for match filtering, be
$\{0,1\}^{128}$; that is, the set of 128-bit binary vectors.  Let
$\mathcal{F}$, the set of possible filters, be $\mathcal{V} \times
\mathbb{Z}$; each filter is a pair $(m,\lambda)$ of a vector $m$ from
$\mathcal{V}$ (called the \emph{mask}) and an integer $\lambda$.  We call
filters of this type \emph{lambda filters}.  Let
$\mathit{check}((m,\lambda),v)=\mathsf{T}$ if and only if strictly more than
$\lambda$ bits are $1$ in the bitwise AND of $m$ and $v$.  Where these
filters come from (the function $\mathit{filt}$) will be discussed later;
for now, note that we can create a match-everything filter by setting
$\lambda=-1$, regardless of the vectors $m$ and $v$; so with an appropriate
definition of $\mathit{filt}$ these definitions
are capable of describing a filtering scheme that is at least correct if not
highly efficient.

The function $\mathit{vec}:\mathcal{E} \rightarrow \mathcal{V}$, which
associates a 128-bit vector with an EIDS tree, is defined as follows.  Let
$T$ be the input tree.  The 128-bit result is divided into four 32-bit
words; call them $v_1$, $v_2$, $v_3$, $v_4$.  A hash function chooses three
distinct bits in $v_1$ to be set to $1$, depending on the head of $T$, or
three bits representing the hash of the empty string if there is no head. 
These bits must be distinct; it is a uniform choice among the
$\binom{32}{3}=4960$ combinations of three out of 32 bits.  Then another
hash function sets three more bits (distinct from each other but not
necessarily from the three representing the head) depending on the arity and
functor of $T$.  Thus, $v_1$ will contain between three and six $1$ bits.

If we are looking for trees that exactly match a specific head at the root,
we can say with certainty that any tree having the desired head will have
three specific bits in its vector equal to $1$, namely the three bits
corresponding to the hash of the head.  A filter $(m,2)$ with $m$ selecting
exactly those three bits will match all such trees---and not many others,
because with only at most six bits of 32 set, the chances are good that at
least one of the three bits will be $0$ on a tree that does not have the
desired head.  This filter foreshadows the more complicated filters we will
use to approximate the full EIDS $\mathit{match}$ operation.

In the case of a nullary tree, the calculation of $\mathit{vec}(T)$ stops at
this point, leaving $v_2=v_3=v_4=0$.  For higher arities, we recursively
compute the vectors for the children and merge them as follows, where
$(w_1,w_2,w_3,w_4)$, $(x_1,x_2,x_3,x_4)$, and $(y_1,y_2,y_3,y_4)$ are the
values of $\mathit{vec}$ for the children in order, split into 32-bit
words, and $|$ represents the bitwise OR operation.
\begin{itemize}
  \item If $T$ is unary, then $v_2=v_3=w_1$ and $v_4=w_2|w_3|w_4$.
  \item If $T$ is binary, then $v_2=w_1$, $v_3=x_1$, and
    $v_4=w_2|w_3|w_4|x_2|x_3|x_4$.
  \item If $T$ is ternary, then $v_2=w_1$, $v_3=y_1$,
    and $v_4=w_2|w_3|w_4|x_1|x_2|x_3|x_4|y_2|y_3|y_4$.
\end{itemize}

\begin{figure}
  \centering
  \begin{tikzpicture}[yscale=1.3,xscale=2]
    \node[anchor=west] at (-4.8,1.0) {$\phantom{v_1 =}
      \texttt{05A0\,B040\,2090\,0420\,03C0\,1040\,8800\,5401}_{16}=
      \mathit{vec}(\text{\eids{<語>⿰言<吾>⿱五口}})$};
    \node[anchor=west] at (-4.8,0) {$v_1 = \texttt{8800\,5401}_{16}$};
    \node[anchor=west] at (-4.8,-0.3) {$\phantom{v_1} =
      1000\,1000\,0000\,0000\,0101\,0100\,0000\,0001\,_2$};
    \node[anchor=west] at (-4.8,-1) {$v_2 = \texttt{03C0\,1040}_{16}$};
    \node[anchor=west] at (-4.8,-1.3) {$\phantom{v_2} =
      0000\,0011\,1100\,0000\,0001\,0000\,0100\,0000\,_2$};
    \node[anchor=west] at (-4.8,-2) {$v_3 = \texttt{2090\,0420}_{16}$};
    \node[anchor=west] at (-4.8,-2.3) {$\phantom{v_3} =
      0010\,0000\,1001\,0000\,0000\,0100\,0010\,0000\,_2$};
    \node[anchor=west] at (-4.8,-3) {$v_4 = \texttt{05A0\,B040}_{16}$};
    \node[anchor=west] at (-4.8,-3.3) {$\phantom{v_4} =
      0000\,0101\,1010\,0000\,1011\,0000\,0100\,0000\,_2$};
    \node at (0,0) {\scalebox{1.5}{\eids{<語>[⿰]}}};
    \node at (-0.8,-1.6) {\scalebox{1.5}{\eids{<言>(;)}}};
    \node at (0.8,-1.6) {\scalebox{1.5}{\eids{<吾>[⿱]}}};
    \draw[>=latex',->,ultra thick] (-0.25,-0.4) -- (-0.55,-1.2);
    \draw[>=latex',->,ultra thick] (0.25,-0.4) -- (0.55,-1.2);
    \node at (0,-3.2) {\scalebox{1.5}{\eids{<五>(;)}}};
    \node at (1.6,-3.2) {\scalebox{1.5}{\eids{<口>(;)}}};
    \draw[>=latex',->,ultra thick] (0.55,-2.0) -- (0.25,-2.8);
    \draw[>=latex',->,ultra thick] (1.05,-2.0) -- (1.35,-2.8);
    \draw[>=latex',->,dashed,thick] (-0.3,0.2)
      to[bend right=30] node[pos=0.3,above] {\small 1, 15, 28} (-1.4,-0.1);
    \draw[>=latex',->,dashed,thick] (0.3,-0.25)
      to[bend left=30] node[pos=0.6,above] {\small 11, 13, 32} (-1.4,-0.4);
    \draw[>=latex',->,dashed,thick] (-1.0,-1.4)
      to[bend right=30] node[pos=0.2,above] {\small ~~~~~~23, 24, 26}
        (-1.5,-1.1);
    \draw[>=latex',->,dashed,thick] (-0.7,-1.85)
      to[bend left=30] node[pos=0.5,above] {\small 7, 13, 25} (-2.4,-1.5);
    \draw[>=latex',->,dashed,thick] (0.5,-1.4)
      to[bend right=10] node[pos=0.2,above] {\small 6, 11, 28} (-1.4,-2.1);
    \draw[>=latex',->,dashed,thick] (1.1,-1.85)
      to[bend left=20] node[pos=0.5,above] {\small 6, 25, 30} (-1.4,-2.5);
    \draw[>=latex',->,dashed,thick] (-0.3,-3.0)
      to[bend right=30] node[pos=0.8,above] {\small 15, 16, 22} (-2.4,-3.1);
    \draw[>=latex',->,dashed,thick] (0.2,-3.45)
      to[bend left=30] node[pos=0.3,above] {\small 7, 13, 25} (-2.4,-3.4);
    \draw[>=latex',->,dashed,thick] (1.3,-3.0)
      to[bend right=30] node[pos=0.2,above] {\small 16, 24, 27} (-1.4,-3.1);
    \draw[>=latex',->,dashed,thick] (1.8,-3.45)
      to[bend left=30] node[pos=0.2,above] {\small 7, 13, 25} (-1.4,-3.4);
  \end{tikzpicture}
  \caption{Calculating the $\mathit{vec}$ function.}
  \label{fig:haystack-bv-example}
\end{figure}
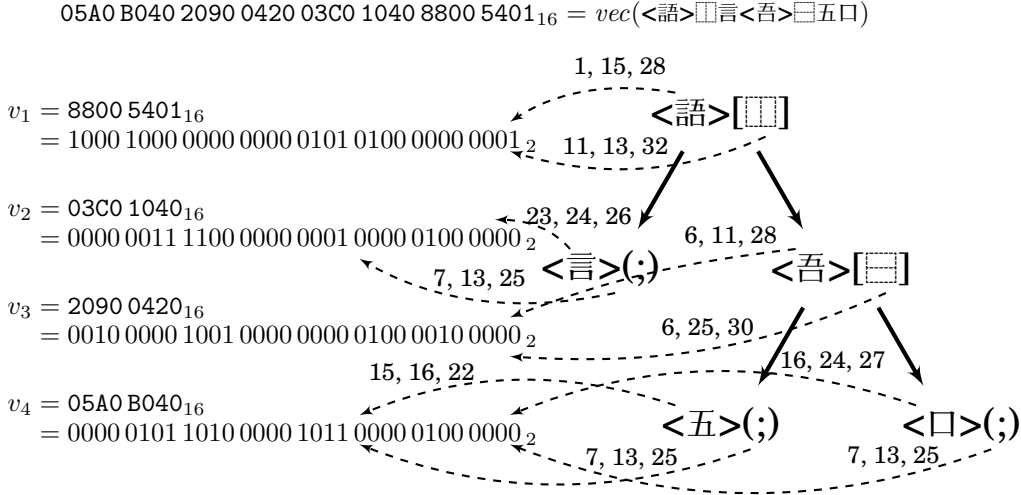

An intuitive description of the $\mathit{vec}$ calculation is that $v_1$
represents the head and functor of the root of $T$; $v_2$ represents the
head and functor of the first child of the root; $v_3$ represents the head
and functor of the \emph{last} child (which could be first, second, or third
depending on the arity); and $v_4$ represents any other descendants of the
root, including a middle child if any, grandchildren, and deeper
descendants.  Figure~\ref{fig:haystack-bv-example} illustrates the bit
vector calculation.  The last word, $v_4$, will tend to be dense in $1$ bits
for large trees, because it represents the bitwise OR of Bloom filters for
an unlimited number of nodes in the tree.  But as long as it contains a few
zero bits, it can be of some use in ruling out matches for queries that
touch on those bits.  The EIDS trees representing Han characters are often
shallow (one or two, rarely more, layers of descendants below the root), so
that accurately representing the root and two of its children is often
enough for useful filtering; and with a sparse 32-bit word for each of the
root and first and last children, we can hope for a high rejection rate on
any matches to those nodes.

Figure~\ref{fig:haystack-bv-example} illustrates the bit vector calculation
for a dictionary entry, with values in hexadecimal and binary notation as
indicated by the subscripts.  Words in the vector, and bits in the words,
are indexed in one-based little-endian order; the least significant bit of
the vector is bit 1 of $v_1$.  Each node in the tree selects three bits with
its head (or lack of a head) and three bits with its functor/arity pair, as
shown by the indices on the dashed arrows; these bits are set to $1$ in a
word selected by the location of the node in the tree.  All three of the
nullary nodes with semicolon functors select the bit combination 7, 13, 25
in their respective words.  In the case of a unary root (rare in Han
character dictionary entries, but they occur in search patterns), the single
child would set bits to $1$ in both $v_2$ and $v_3$.  Both of the two
grandchildren of the root select $v_4$, so it ends up with a greater density
of $1$ bits than the other words.

We mentioned that a filter $(m,2)$ with $m$ containing three $1$ bits in the
first 32-bit word of $m$, chosen by hashing to encode the head of the root
of an EIDS tree, is a filter for the query that would match exactly trees
having that head.  It matches the vectors of all those trees, and few
others.  Lambda filters are constructed the same way for matching the state
of having no head at the root, and for matching a functor/arity pair. 
Starting from these \emph{atomic} lambda filters, the filter calculus for
lambda filters combines them to produce lambda filters that approximate more
complicated queries.

Let $f_1=(m_1,\lambda_1)$ and $f_2=(m_2,\lambda_2)$ be lambda filters.  Let
$f_3=(m_3,\lambda_3)$ where $m_3=m_1|m_2$ and $\lambda_3=min
\{\lambda_1,\lambda_2\}$.  Any vector that matches $f_1$ or $f_2$ must
contain more than $\lambda_1$ of the $1$ bits specified by $m_1$, or more
than $\lambda_2$ of the $1$ bits specified by $m_2$; so it necessarily
contains more than the minimum of those ($\lambda_3$) in the union of those
bit positions ($m_3$).  Therefore the lambda filter $f_3$ can be used to
approximate the Boolean OR of the functions approximated by $f_1$ and $f_2$,
with correct results.  As we combine more and more filters with OR, the
precision of the result will tend to decrease, because $\lambda$ tends to
decrease while the number of $1$ bits in $m$ tends to increase.  Looking for
fewer bits to be $1$ among a larger number of bits selected by the mask
increases the chances of a false positive.  However, in a simple query the
number of consecutive OR operations may be small enough for this loss of
precision not to be a problem.

If we know that a vector $v$ contains more than $\lambda_1$ of the $1$ bits
in $m_1$ \emph{and} more than $\lambda_2$ of the $1$ bits in $m_2$, then by
counting the $1$ bits included in $m_1$ but not in $m_2$, in $m_2$ but not
in $m_1$, and in the bitwise AND of $m_1$ and $m_2$, we can derive a set of
linear inequalities on the number of bits $v$ contains in each of those
categories.  From those inequalities (not shown here; the derivation is
lengthy, although elementary) we can find seven different lambda filters
that will necessarily match $v$, corresponding to the seven nontrivial
subsets of the bit categories $\{$ in $m_1$ but not $m_2$, in $m_2$ but not
$m_1$, in both $m_1$ and $m_2$\}.

Any of these seven filters would produce correct results if used as the
lambda filter for the Boolean AND of $(m_1,\lambda_1)$ and
$(m_2,\lambda_2)$; but depending on the nature of the input filters and the
distribution of vectors in the database, some of the possible results may be
much better than others for efficiency.  Some may be trivial
match-everything filters.  IDSgrep's implementation uses an heuristic rule
to attempt to choose the combination that will give the most precise filter:
it will include each of the three bit categories if more than one third of
the bits in that category are required to be $1$, unless no categories meet
that criterion, in which case it will use any categories that require at
least one bit to be $1$.  This rule is intended to maximize the $\lambda$
value of the resulting filter, which should help precision, while avoiding
the use of very dense masks, which could be expected to harm precision. 
Whether it is a good rule is difficult to evaluate in isolation, but bears
on the experimental results for the overall filtering scheme.

One more filter calculus operation is necessary to do basic EIDS matching:
we must be able to compute a lambda filter for matching a child, given that
we have a lambda filter for matching at the root.  Here something similar to
the Boolean AND calculation applies.  If $T$ and $U$ are EIDS trees, $T$ is
the first child of $U$, and more than $\lambda$ selected bits in the first
word of $\mathit{vec}(T)$ are $1$, then more than $\lambda$ of the
corresponding bits in the \emph{second} word of $\mathit{vec}(U)$ are $1$,
because those are the same bits by definition of $\mathit{vec}$.  That is
fine as long as our lambda filters examine only the first word.  However,
there can be multiple bits (as many as three) in $\mathit{vec}(T)$ that are
combined with bitwise OR to form a single bit in $\mathit{vec}(U)$, and if
two bits in $\mathit{vec}(T)$ are $1$ but collide in this way, we only know
of one bit in $\mathit{vec}(U)$ guaranteed to be $1$.  By examining the bits
in the mask $m_1$ of a lambda filter $f_1=(m_1,\lambda_1)$ and determining
the worst-case number of collisions (this derivation, like that for AND, is
lengthy but elementary), it is possible to construct a filter
$f_2=(m_2,\lambda_2)$ that will necessarily match any tree whose first child
would be matched by $f_1$.  Similar constructions exist for second and
third children.  Applying this construction repeatedly can produce a filter
for matching any chosen descendant of the root.

Now we have the necessary filter calculus operations to do basic EIDS
matching.  Recall that under basic matching, that is, when the functor and
arity of the needle $N$ do not correspond to a special matching operator,
$\mathit{match}(N,H)=\mathsf{T}$ if and only if either $N$ and $H$ both have
heads and those heads are identical, or $N$ and $H$ do not both have heads
but they have the same functor and arity and all corresponding children
match recursively.  Each operation in that logical statement corresponds to
a filter calculus operation.  Starting with the atomic lambda filters for
matching heads, head absence, and functor/arity, we can use OR,
AND, and the ``match child'' transformation to calculate a lambda filter
$\mathit{filt}(N)$ with the desired property that
$\mathit{check}(\mathit{filt}(N),\mathit{vec}(H))=\mathsf{T}$ whenever $N$
and $H$ match under basic matching, and not often otherwise.

It takes a little more work to handle the special matching operators of
IDSgrep.  When calculating a filter for a query, IDSgrep follows the same
logic as in the definition of the $\mathit{match}$ function: check whether
the functor and arity of the root of the query EIDS tree describe a special
matching operator, follow rules specific to the operator if one is
recognized, and apply basic matching otherwise.  The Boolean OR and AND
operations \eids{[|]} and \eids{[\&]} are handled by straightforward
application of the OR and AND rules already defined by the filter calculus.

The Boolean NOT operator \eids{.!.} is more difficult.  Recall that filters,
by definition, must match when the full EIDS matching query would match, but
need not make any guarantees about whether they will match or not in other
cases.  The match-everything filter is an acceptable filter for every
query, and any filter is acceptable for the match-nothing query.  As a
result, we cannot by examining a filter determine with certainty any
circumstances under which the corresponding EIDS matching query would
\emph{not} match.  Any case of the filter matching might be a false positive
for which the full query would not match, and would become a forbidden false
negative if we attempted to invert it.  Therefore, if we attempt to evaluate
the \eids{.!.} operator in pure filter calculus where given a filter for a
query $x$ we must find a filter for \eids{.!.}$x$, the only correct result
will be a match-everything filter, regardless of $x$; and further filter
calculus operations using that match-everything filter will tend to yield
imprecise results, harming the overall outcome of the filtering.

IDSgrep processes the \eids{.!.} operator by temporarily breaking out of the
filter calculus to apply Boolean algebra to the underlying EIDS
matching queries, which contain more useful information than their lambda
filters.  It applies double negation (NOT NOT $x$ equivalent to $x$), de
Morgan's theorem (NOT ($x$ OR $y$) equivalent to (NOT $x$) AND (NOT $y$)),
and recognizes the special cases of the match-everything and match-nothing
queries \eids{(?)} and \eids{.!.(?)}.  Whenever it would calculate the lambda
filter of a query with the \eids{.!.} operator at the root, it instead applies
any of these rules that it can, to push the negation further down in the
tree.  Doing so may sometimes eliminate the negation entirely, but even if
it cannot be eliminated, negation further down in the tree will tend to
cause less harm to the precision of the final filter.  If the
negation cannot be removed or postponed, only then does IDSgrep resort
to returning the match-everything filter.

Only a few other special operators are handled by the lambda filtering
scheme.  The filter for the literal-match operator \eids{.=.} is just the
filter for its child under basic matching, ignoring any special meaning of
the child's functor and echoing the definition of \eids{.=.}.  The
unordered-match operator \eids{.*.} is expanded into an equivalent
construction using Boolean OR on all permutations of children, before
calculating the lambda filter on the expansion.  Similarly, the
match-anywhere operator \eids{...} is expanded into an equivalent OR of four
queries:  one each for matching at the root, first child, last child, and
all other descendants.  These cases correspond neatly to the four 32-bit
words in the bit vector.

One can imagine a similar expansion of an associative query using \eids{.@.}
into an equivalent query without \eids{.@.}, but such a construction would
suffer from a combinatorial explosion, containing one subquery for each way
of parenthesizing the original, all combined with Boolean OR.  IDSgrep
avoids this possibility by just using the match-everything filter for
\eids{.@.} queries, giving behaviour that is at least correct and not
significantly worse than no filtering at all.  Similarly, the user-predicate
and regular-expression matching operators, which escape to other matching
functions that defy algebraic analysis, are always assigned match-everything
lambda filters.

By applying these rules, IDSgrep can calculate a lambda filter for any
query, having the desired properties of no false negatives and reasonably
few false positives.   That is the first layer of filtering, used to avoid
both explicit calculation of the $\mathit{match}$ function and evaluation of
the somewhat more expensive second layer of filtering, which is described
next.


\subsection{BDD filters}

The precision of lambda filtering is limited by the implicit requirement of
the filter calculus that the result of an operation on lambda
filters must itself be a lambda filter.  If we consider
$\mathit{check}(\mathit{filt}(N),\cdot)$ as a Boolean function on bit
vectors, it may well be that the function we would like it to compute is not
one that can be well-modelled by a lambda filter.  For instance, consider
the lambda filters $(0101,1)$ and $(1010,1)$ on four-bit vectors.  Those
match vectors of the form $x1y1$ and $1x1y$ respectively; four vectors each,
with one vector matched by both filters. 
But the most precise lambda filter for the OR of these two filters is
$(1111,1)$, which matches 11 of the 16 possible four-bit vectors, including
four that would not have been matched by either input to the OR.  If these
filters already represented compromises to the hope of closely approximating
a tree match, the result of the OR is likely to be a much worse compromise. 
We might hope to do better with a more powerful set of Boolean functions
applied to the same bit vectors, and a filter calculus that loses less
precision in its operations.

IDSgrep's \emph{BDD filters} address that hope.  They are named for the
\emph{binary decision diagram} (BDD), which is a well-known data structure
for representing Boolean functions of bit vectors.  The data structure is
well described in standard references~\cite{Knuth:BDD} and we do not explain
its inner workings beyond IDSgrep's perspective.  IDSgrep uses a third-party
open-source BDD library named BuDDy~\cite{BuDDy} as a black box
implementation of BDDs.

BuDDy provides Boolean functions as objects for the software to manipulate,
with operations ranging from simple, like ``compute the function that is the
Boolean OR of these two functions,'' to much more complicated, like ``count
the number of distinct input vectors on which this function is true.'' Some
of these operations are NP-hard and cannot be performed in reasonable time
in the worst case; but the algorithms and the implementation
include many optimizations for the cases expected in practice.

IDSgrep applies BDDs directly to the match filtering problem.  Let
$\mathcal{V}$, the set of vectors for match filtering, be $\{0,1\}^{128}$,
the 128-bit binary vectors just as in lambda filtering.  Similarly, let
$\mathit{vec}$ be the same function used in the lambda filtering of the
previous section.  BDD filtering operates on the same vectors.  It differs
in the definition of $\mathcal{F}$, the set of filters: here, $\mathcal{F}$
is the set of all \emph{monotonic} Boolean functions on 128-bit binary
vectors.  Monotonic Boolean functions of binary vectors are those where, if
the function's value for a given input is true, changing a $0$ bit in the
input to a $1$ can never cause the function's value to change to false. 
This requirement limits the complexity of the functions somewhat, but
$\mathcal{F}$ remains a huge set, and we will later apply a further
constraint on the complexity of the functions that will actually be used. 
Elements of $\mathcal{F}$ are represented by binary decision diagrams, and
the $\mathit{check}$ function on a BDD and a vector simply evaluates the
function that the BDD represents, using the vector as input.

The calculus of BDD filters starts with atomic filters and applies
operations to create filters for arbitrary EIDS matching queries, much like
the calculus of lambda filters.  Just as with lambda filters, when a query
has a given head at the root, there are three bits in the first word of its
vector that must be $1$.  It is easy to create a BDD for the function true
if and only if all those bits are $1$, and that is the atomic BDD filter for
matching that head value.  So far, it represents the same function that the
equivalent lambda filter would represent.  Similar atomic BDDs are easy to
define for matching the absence of a head at the root, and any given
functor/arity pair.

Boolean OR and AND use the relevant BDD operations directly.  Here is the
first significant difference from lambda filtering: whereas the lambda
filter for the OR of two lambda filters may also match on some vectors that
would not have been matched by either input (a loss of precision), the OR of
two BDDs represents \emph{exactly} the function that is true if and only if
at least one of the input functions is true; and the BDD for AND is,
similarly, an exact representation of that operation.  There is no loss of
precision in these simplest BDD filter calculus operations.

Matching a child may involve some loss of precision.  Suppose for some EIDS
tree $T$ which is the first child of the EIDS tree $U$, the four 32-bit
words of $\mathit{vec}(T)$ are $(v_1,v_2,v_3,v_4)$ and the four 32-bits
words of $\mathit{vec}(U)$ are $(w_1,w_2,w_3,w_4)$.  Consider bit 1 of
$w_4$.  If it is $1$, that could be because bit 1 of any of $v_2$, $v_3$, or
$v_4$ might have been $1$, or as a result of other bits coming from other
children of $U$.  The filter has no way of determining which of those
possibilities might be true just by examining $\mathit{vec}(V)$, except that
if the bit is $1$ then \emph{at least one} of the bits that contributed to
it must have been $1$.  However, recall that we required BDD filters to
represent monotonic functions.  If bit 1 of $w_4$ is $1$, then the filter
can safely assume that \emph{all} of the bits that were combined with OR to
generate that bit---that is, all of $v_2$, $v_3$, $v_4$, and the similar
bits from other children of $U$---were $1$.  Then the filter will
necessarily match in all the cases where it is required to match, as well
as possibly some other cases; some precision is lost here.

For each bit in the child's vector, we can write a \emph{rearrangement
function} of the bits in
the parent's vector that describes whether the child bit could possibly be
true.  For instance, bit 1 of $v_1$ could possibly be true if and only if
bit 1 of $w_2$ is true.  The collection of such functions produces a safe
guess at the contents of $\mathit{vec}(T)$ based on $\mathit{vec}(U)$:
a vector guaranteed to contain $1$ at every bit position where
$\mathit{vec}(T)$ contains $1$.  By monotonicity, a BDD filter applied to
the guess will produce a usable approximation of the same filter's results
applied to the actual value of $\mathit{vec}(T)$, only with a possible loss
of precision.  Therefore to match a pattern as the first child of the root,
IDSgrep builds a BDD filter to match the same pattern at the root,
composes it with the collection of bit rearrangement functions, and the
result is a filter to match the same pattern as the first child instead of
at the root.  A
similar construction is used to match second or third children.

Applying these operations to the atomic filters, as in lambda filtering,
gives BDD filters for basic EIDS matching.  Filters for special matching
operators are also constructed using similar techniques to those used for
lambda filtering.  Boolean OR and AND, and the literal match operator
\eids{.=.}, are straightforward.  Boolean NOT is
handled by examining the underlying EIDS-match queries and applying Boolean
algebra, as in lambda filtering, with a match-everything BDD filter used as a
fallback where necessary.  Unordered match \eids{.*.} is expanded into a
Boolean OR of the matched permutations, and match-anywhere \eids{...} into
an OR of four expressions for matching at the root, as first child, as last
child, or as any other descendant.  Finally, the \eids{.@.}, \eids{./.}, and
\eids{.\#.} operators are assigned match-everything BDD filters, as in the
lambda filtering case.

One significant issue remains:  the complexity of the calculated BDD
filters.  The BuDDy library is quite efficient, containing most of the usual
optimizations expected of a BDD library, and it has a minor advantage over
other libraries for IDSgrep's purposes because it does \emph{not} include
a popular optimization called \emph{negated edges}.  Because IDSgrep operates
only on monotonic functions, negated edges would never actually be used, and
leaving the fields for storing them out of the data structure improves the
constant factors.  But even with good constants, any BDD library must
repeatedly solve NP-hard problems to maintain the data structure, and there
is a potential for both time and space requirements to become exponential. 
We could imagine that a pathological query would cause IDSgrep to spend so
much time in the per-query preprocessing, building the BDD filter, as to
outweigh any possible advantage in the per-entry scanning.

IDSgrep addresses that concern by enforcing a constraint on the complexity
of any BDD returned by filter calculus operations.  Recall that adding false
positives to a filter will never cause incorrect results from the overall
filtering and matching algorithm; it will only reduce efficiency by
requiring more full EIDS tree matches.  We can always change a BDD filter to
one that returns the possible-match result on more vectors, as long as we do
not cause it to stop returning possible-match on any vectors for which it
already does so.  Furthermore, the BuDDy library can provide an estimate of
the cost of a BDD in time and space, in the form of a count of the nodes in
an internal data structure.  With these facts in mind, IDSgrep has a simple
way of avoiding excessive resource consumption in BDD filter calculus
operations:  after each operation, it checks whether the result is too
complicated, and if so, replaces it with something simpler but still
correct.

In more detail, after each BDD filter calculus operation IDSgrep checks the
number of nodes in the BDD.  If the BDD contains more than 1000 nodes, it
applies \emph{existential quantification} to one of the bits in the input. 
For a given bit $b_i$ in the input, the existential quantification of the
BDD of a function $f_1$ with respect to $b_i$ is a new BDD of a function
$f_2$ that is true if and only if some value of $b_i$ (either $1$ or $0$)
exists such that $f_1$ is true, on input vectors otherwise identical.  For
monotonic functions this can be thought of as forcing the value of $b_i$ to
be $1$.  The bit $b_i$ ceases to be in the support of the function (that is,
$f_2$ no longer depends on the value of $b_i$) and so the BDD of $f_2$
contains no internal nodes mentioning that bit.

Applying existential quantification to every bit of the input in turn would
eventually yield a one-node BDD that is identically true or identically
false regardless of the input; therefore, it is always possible to reduce
the size of the BDD to 1000 nodes or less by applying a sufficient number of
quantification operations.  IDSgrep does so, choosing bits in decreasing
order of index from the most significant in $v_4$ down to the least
significant in $v_1$ until the node count is less than or equal to 1000. 
The result is a BDD filter that matches at least all vectors matched by the
unconstrained filter, but with a limited complexity.  Enforcing this
constraint after every filter calculus operation prevents the next filter
calculus operation from taking excessive amounts of time or space.  Some
precision may be lost, but only on complicated queries for which match
filtering provides little advantage to begin with.


\subsection{Match memoization}

Straightforward recursive descent evaluation of the IDSgrep matching
function takes exponential time in the worst case.  The definition of
$\mathit{match}$ recurses more than once into its children in the cases of
the \eids{...} operator (each subtree of the haystack against the needle)
and the \eids{.*.} operator (the haystack against as many as six
permutations of the needle).  A matching pattern with many nested instances
of these may take a very long time to evaluate.

However, the straightforward recursive descent algorithm lends itself to
dynamic programming via memoization.  The needle and haystack each contain a
linear number of subtrees, and each pair of subtrees deterministically does
or does not match.  We can store and re-use the result of
$\mathit{match}(N,H)$ for each pair $(N,H)$ in a table of size $O(n^2)$. 

Computing the $\mathit{match}$ function given the table entries for all
subtrees of its arguments is a linear-time operation in the worst case
implemented within IDSgrep, which is the \eids{.@.} operator; that operator
potentially requires comparing node lists of linear length.  IDSgrep stores
strings using a hashed symbol table for constant-time equality tests on
strings, so \eids{.@.} can be implemented in $O(n)$ time, plus recursion
into the subtrees.  The other operators are constant-time after recursion is
paid for.  Multiplying $O(n^2)$ subproblems with $O(n)$ time per subproblem
gives $O(n^3)$ time overall.  This analysis excludes the \eids{./.}
regular-expression matching operator.  That operator connects IDSgrep to the
external PCRE library, which does not offer time guarantees; but $O(n^2)$
remains as a bound on the number of calls IDSgrep makes to PCRE.

In the practical implementation, on commonly-occurring queries, match
memoization is rarely beneficial.  Users seldom construct queries with more
than one or two instances of \eids{...} or \eids{.*.}, rarely nesting them
even then.  The additional constant factors associated with hashing before
and after each node-to-node matching test, increased memory working set size
resulting from random accesses to the hash table, and so on, are
considerable.  But to guard against pathological or malicious queries, the
IDSgrep utility implements memoization conditional on the matching pattern. 
When the matching pattern includes more than two instances of \eids{...} or
\eids{.*.}, IDSgrep will memoize $\mathit{match}$, giving a $O(n^3)$ time
bound while still avoiding the overhead of maintaining the hash table in the
usual case of simpler queries.


\section{Experimental evaluation}
\label{sec:evaluation}

This section presents experimental evaluation of IDSgrep version 0.5.1, with
BuDDy 2.4 and dictionaries from CJKVI as supplied by the IDSgrep
distribution; CHISE-IDS 0.25; the September 1, 2013
released version of KanjiVG; and Tsukurimashou 0.8.  PCRE 8.31 was available
on the experimental computer, but only used in the present experiments for
running IDSgrep's own test suite to verify that the software had been
compiled and installed correctly.  Speed results are user CPU time on one
core of an Intel Core i5-2400S desktop computer with 4G of RAM and a 2.5GHz
clock, running 64-bit Scientific Linux 6.5 with Linux kernel version
2.6.32-358.18.1.el6.  IDSgrep was compiled in the default configuration
selected by its build scripts; on this system that invoked the GCC 4.4.7
compiler with ``-O2'' optimization.


\subsection{Match filtering}

The main experimental question of interest here was how the
algorithmic enhancements (both kinds of match filtering, and match
memoization) affect query speed.
The speed test queries were chosen to be similar to those users typically
make in practice, and to exercise the relevant features of the query
language.  There was an emphasis on queries involving
wildcards and Boolean logic, which are more challenging to search
algorithms.  Some queries returning no hits, and some returning large
numbers of hits, were tested.  However, artificial pathological cases that
users would not be expected to create in actual use were not included in the
main speed evaluation.

We started with the 160 Grade Two J\={o}y\={o} Kanji characters as taught in
the Japanese school system, and found their entries in the CJKVI
Japanese-language character structure dictionary generated by the IDSgrep
installer.  That dictionary excludes characters with no breakdown into
smaller components, according to its own rules for determining what
qualifies as an atomic component; other dictionaries do have entries for
some of the characters that CJKVI excludes.  For 144 of the Grade Two
characters, CJKVI provided an entry; and for each of those we removed
all heads from the EIDS tree except at the leaves, to create a tree that
might be further modified to form a test query.  For instance, from the
dictionary entry \eids{【数】⿰<娄>⿱米女攵}, removing the non-leaf heads
gave \eids{⿰⿱米女攵}.

The test query set contained 1642 queries and was constructed as follows:
\begin{itemize}
   \item All 160 Grade Two kanji as single characters for head-to-head
     matching.
   \item Match-anywhere applied to each of the 160 Grade Two kanji.
   \item The 144 dictionary entries with heads removed.
   \item The 144 headless dictionary queries with each leaf in turn replaced
     by the wildcard \eids{(?)}.  For instance, \eids{⿰⿱米女攵} generated
     \eids{⿰⿱?女攵}, \eids{⿰⿱米?攵}, and \eids{⿰⿱米女?}.  This process
     created 536 queries, reduced to 524 by removing duplicates.
   \item Unordered-match applied to the root of each of the 144 headless
     dictionary entries, for instance \eids{*⿰⿱米女攵} from
     \eids{⿰⿱米女攵}.
   \item For all headless dictionary entries that included the necessary
     structure for associative match to be meaningful, such as
     \eids{⿰日⿱⿱十一寸}, the same tree with associative match inserted,
     such as \eids{⿰日@⿱⿱十一寸}; there were 53 of these, including three
     where it was possible to apply \eids{@} to two different associative
     structures in the same tree.
   \item For all headless dictionary entries with binary roots, the
     same tree with the root replaced by the Boolean OR operator, for
     instance \eids{|口儿} from \eids{⿱口儿}.  There were 137 of these.  The
     seven headless dictionary entries without binary roots all had
     \eids{⿳} as root functor.
   \item For each $x$ chosen from among the 160 Grade Two kanji,
     the queries \eids{\&...}$x$\eids{...日} and
     \eids{\&...}$x$\eids{!...日}.
     That makes 320 queries, intended to test Boolean AND and NOT with
     match-anywhere in usage patterns similar to multi-radical search; since
     \eids{日} occurs within some of the Grade Two kanji, some of these
     queries will necessarily return no results.
\end{itemize}

The literal-match, regular-expression, and user-defined predicate operators,
which exist for special purposes not directly relevant to structural query
of Han characters, were excluded from the test query list.  The IDSgrep
package includes a test suite of its own with a similar selection of queries
to ours, but it is based on the Grade One characters and their dictionary
entries in KanjiVG, with some manually-constructed pathological queries
included to test correctness of the implementation rather than speed in
realistic use.

Four dictionaries of character decompositions were used for the speed test:
CJKVI (Japanese version, supplied in the IDSgrep 0.5.1 package) with 74361
entries totalling 4461882 bytes; CHISE IDS version 0.25, with 133606 entries
totalling 5555303 bytes; the KanjiVG release of September 1, 2013, with 6666
entries totalling 175257 bytes; and Tsukurimashou 0.8, with 2655 entries
totalling 106021 bytes.  This makes a total of 217288 dictionary entries. 
Although CHISE IDS supplies more than half the entries, the other
dictionaries often use different structural descriptions of
frequently-occurring characters and components, and so they add some
diversity in the trees to be searched.

The IDSgrep 0.5.1 installer is also capable of building a dictionary from
the EDICT2 file~\cite{EDICT2}, but that was not included in the present
experiment because it is a dictionary of word meanings and pronunciations,
intended to be searched primarily with PCRE.  Since it contains many large
entries that would tend to be skipped by the test queries aimed at single
characters, its inclusion in the timing results would tend to overstate the
advantages of bit filtering in the character dictionary applications studied
here.

Table~\ref{tab:overview} summarizes the test queries, test dictionaries, and
numbers of hits returned (final tree matches, which are the same regardless
of filtering).  The mean number of hits per query was 101.72.  The top three
queries by number of hits were \eids{...心}, \eids{\&...心!...日}, and
\eids{...止}, returning 5353, 5152, and 4074 hits respectively.  There were
67 queries that returned no hits, and 560 that returned one hit each.  Note
that the total hit counts for Boolean AND and match-anywhere queries are the
same because of the design of the test query set: each
match-anywhere test query corresponds to a pair of Boolean AND test queries
whose results are disjoint and when unified are the same hits
returned by the match-anywhere query.

\begin{table}
  \tbl{Test queries, test dictionaries, and tree-match hit counts.}{
  \begin{tabular}{lr|rrrr|r}
    & queries & CJKVI-J & CHISE & KanjiVG & Tsuku. & TOTAL \\
    dictionary size & & 74361 & 133606 & 6666 & 2655 & 217288 \\ \hline
    Grade Two kanji & 160 &
      144 & 123 & 160 & 320 & 747 \\
    match-anywhere Gr.\ 2 & 160 &
      28757 & 37394 & 1903 & 634 & 68688 \\
    headless dictionary entries & 144 &
      152 & 67 & 18 & 16 & 253 \\
    wildcard leaves & 524 &
      16237 & 9998 & 1212 & 357 & 27804 \\
    unordered match & 144 &
      157 & 67 & 18 & 16 & 258 \\
    associative match & 53 &
      54 & 9 & 0 & 2 & 65 \\
    Boolean OR & 137 &
      145 & 31 & 129 & 213 & 518 \\
    Boolean AND & 320 &
      28757 & 37394 & 1903 & 634 & 68688 \\ \hline
    TOTAL & 1642 &
      74403 & 85083 & 5343 & 2192 & 167021
  \end{tabular}
  }
  \label{tab:overview}
\end{table}

We ran 20 loops of the 1642 test queries against the 217288 entries of the
test dictionaries in each of four treatments: the default IDSgrep
configuration (which includes both lambda and BDD filtering), with lambda
filtering alone, with BDD filtering alone, and with no match filtering. 
Filtering treatments were selected using IDSgrep's built-in command-line
options, and times were collected using its statistics option.  The results
are in Table~\ref{tab:speed-test}.  Hit percentages refer to the input of
each filtering or matching layer; for instance, when both filters were used
the 30980198 BDD hits per loop represented 19.8\% of the 156617732 trees
that had already passed the lambda filter.  All the results shown are per
loop of $1642\times 217288 = 356786896$ matching tests.  The means and
sample standard deviations across the 20 loops are shown for the times,
which are measured in user CPU seconds using the Linux \texttt{getrusage}
system call.  Because the filtering and tree-match algorithms are
deterministic, the filter hit counts are the same for all loops of each
condition.

\begin{table}
  \tbl{Timing and hit count results for the filtering layers.}{
  \begin{tabular}{l|rrrrrr|rr}
    filters & \multicolumn{2}{c}{$\lambda$ hits} &
      \multicolumn{2}{c}{BDD hits} &
      \multicolumn{2}{c|}{tree hits} &
      mean & st.\ dev. \\ \hline
    both &
      156617732 & (43.9\%) & 30980198 & (19.8\%) & 167021 & (0.54\%) &
      173.69 & 0.68 \\
    BDD & & 
      & 30980206 & (8.7\%) & 167021 & (0.54\%) &
      178.02 & 0.74 \\
    $\lambda$ &
      156617732 & (43.9\%) & & & 167021 & (0.11\%) &
      533.60 & 0.24 \\
    none &
      & & & & 167021 & (0.05\%) & 1024.85 & 0.22
  \end{tabular}
  }
  \label{tab:speed-test}
\end{table}

The times for both filter types, and BDD only, seem close enough for a
statistical test to be appropriate.  One-way ANOVA applied to the sample
mean times gives $F(3,76)>11.4\times 10^6$, $p< 0.0001$, so we reject the
null hypothesis that the means are the same for the four filtering
treatments.  The Tukey HSD (Honestly Significant Difference) test applied to
the pairwise differences gives $\mathit{HSD}(0.01)=0.54$, less than the
difference between any pair of sample mean times; all the pairwise
differences are statistically significant with $p<0.01$.


\subsection{Memoization}

To check the effects of memoization, we compiled a modified version of the
IDSgrep command-line utility in which the test for whether to use
memoization (a function called \texttt{check\_memoization}) was disabled. 
The configuration was otherwise default; in particular, both layers of bit
vector filtering were active.
We then ran queries against the combined test dictionaries for the character
component \eids{寺} nested inside $k$ match-anywhere operators, for $k$ from
1 to 10.  These queries all return the same results, but would be expected
to become slower as $k$ increases.

\begin{figure}
  \includegraphics[scale=1.1]{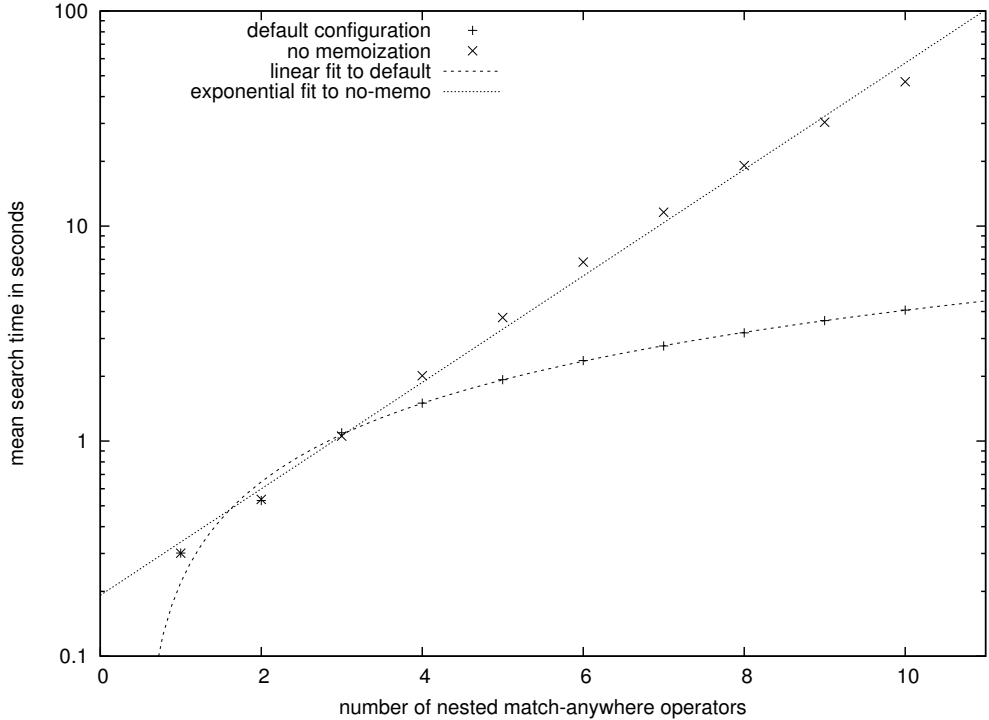}
  \caption{Query times for nested match-anywhere
     with and without memoization.}
  \label{fig:memo-times}
\end{figure}

Table~\ref{tab:memo-times} and Figure~\ref{fig:memo-times} show the time in
seconds per query for each value of $k$, with sample mean and standard
deviation for 100 trials.  Also shown in the figure is a linear function fit
by least squares to the default-configuration query times, and an
exponential function fit to the no-memoization times (by least squares
fitting a line to the logarithms of the data, to avoid overemphasis on the
larger numbers).

\begin{table}
  \tbl{Query times for nested match-anywhere with and without memoization.}{
  \begin{tabular}{l|rr|rr}
    & \multicolumn{2}{c|}{default config}
      & \multicolumn{2}{c}{no memoization} \\
    $k$ & mean & st.\ dev. & mean & st.\ dev \\ \hline
    1 & 0.302 & 0.009 & 0.301 & 0.009 \\
    2 & 0.529 & 0.015 & 0.536 & 0.022 \\
    3 & 1.094 & 0.015 & 1.054 & 0.029 \\
    4 & 1.501 & 0.014 & 2.014 & 0.046 \\
    5 & 1.929 & 0.046 & 3.754 & 0.058 \\
    6 & 2.369 & 0.056 & 6.802 & 0.139 \\
    7 & 2.769 & 0.071 & 11.584 & 0.210 \\
    8 & 3.188 & 0.056 & 19.119 & 0.315 \\
    9 & 3.634 & 0.061 & 30.386 & 0.161 \\
    10 & 4.068 & 0.059 & 46.911 & 0.219
  \end{tabular}
  }
  \label{tab:memo-times}
\end{table}


\subsection{Comparison to other software}

Because IDSgrep is currently the unique implementation of its own query
language, and few similar query languages exist, it is difficult to
meaningfully compare it to other software packages.  We tested two that
might be used for similar purposes to IDSgrep: GNU grep~\cite{grep} and
Tregex~\cite{Levy:Tregex}.

GNU grep is a popular version of the standard command-line grep
utility.  Its basic function is to take a text file as input and pass
through to the output all lines that match a query pattern---much as IDSgrep
does for EIDS trees.  The query patterns for grep are usually described as
\emph{regular expressions}, and regular expressions as such cannot be used
to recognize non-regular languages (the language of balanced parentheses
being the classical example).  EIDS and Unicode IDS, as context-free
non-regular languages, are not well suited to sophisticated queries with
grep.  However, GNU grep, like many recent implementations of grep, supports
back-references and other extensions of regular expression syntax that
allow it to recognize a limited class of non-regular languages.  As one of
its authors describes in an electronic mailing list
posting~\cite{Haertel:Why}, it is a
heavily optimized implementation of standard DFA-based string
search techniques.

Not all IDSgrep queries can reasonably be translated into grep-like string
search queries, but two important kinds of IDSgrep queries easily can be:
looking up a single character with a query like \eids{語} (in which case the
search should return exactly those dictionary entries that have that
character as head), and looking for a single constituent anywhere in the
entry with a query like \eids{...言}.  We can translate these two queries
into GNU grep queries \eids{【語】} and \eids{言} respectively.  In
IDSgrep's default databases, each entry is one text line, lenticular
brackets (synonymous with ASCII angle brackets in EIDS syntax) occur only in
entry heads, and characters like \eids{言} do not occur in unusual contexts,
so these GNU grep queries return the same entries as the original IDSgrep
queries despite not having technically identical semantics.  Simple Boolean
queries can also be performed easily with grep, by passing the output of one
grep instance through another.

In our test query set, 640 queries are thus of a form that can easily be
processed with GNU grep.  We ran 20 loops of those 640 queries against the
combined test databases, using IDSgrep in its default configuration, IDSgrep
with bit vector filtering turned off (for a possibly fairer comparison to
grep, which uses no pre-computed index), and GNU grep.  The timing results
in user CPU seconds, with means and standard deviations over the sample of
20 loops, are shown in Table~\ref{tab:grep-comparison}.

\begin{table}
  \tbl{Timing comparison between IDSgrep and GNU grep.}{
  \begin{tabular}{l|rr}
    search software & mean & std.\ dev. \\ \hline
    IDSgrep (default) & 145.58 & 1.15 \\
    IDSgrep (no filtering) & 478.58 & 1.53 \\
    GNU grep & 9.46 & 0.06
  \end{tabular}
  }
  \label{tab:grep-comparison}
\end{table}

Careful handling of the test databases was necessary for accurate results. 
The third-party data sources used to generate the dictionary files include
some proportion of syntactically invalid entries (worst in the CHISE-IDS
dictionary, where the IDSgrep installer detects 11746 bad entries).  Most of
these are filtered out during dictionary creation, and not included in our
counts of dictionary entries; and IDSgrep's EIDS parser attempts to
tolerate, but cannot correct, any bad syntax that may remain.  Errors in the
third-party dictionaries are not further analysed here.  There is no gold
standard, and the present work only concerns searching in given data,
whatever its quality.  The usual effect of IDSgrep's error recovery is for
it to see more than one tree on the same input line, because of a missing
operator character causing the parse of the first tree to terminate
prematurely.  Thus, it is possible for IDSgrep to return more matches than
the number of lines in the file.  Since GNU grep is strictly line-based, to
get an accurate count of matching trees from grep is it necessary to give it
input with exactly one EIDS tree on each line.  We processed the input
dictionaries through IDSgrep with a match-everything query and its built-in
\emph{cooked output} mode set to generate a canonical EIDS syntax with one
tree on each line.  The processed dictionary file allowed GNU grep to return
line-based match counts identical to IDSgrep's tree-based match counts for
the same queries.

The crucial disadvantage of GNU grep is that it cannot do the complicated
subtree-matching queries for which IDSgrep is intended.  Stanford
Tregex~\cite{Levy:Tregex} is a more powerful tree-matching program
originating in the computational linguistics community, and one of the
nearest pre-existing equivalents to IDSgrep in terms of expressive power and
application domain.  It is intended for use with parse trees of sentences in
databases like the Penn Treebank~\cite{Marcus:Building}, and it supports a
query language based on describing constraints between nodes.  The available
constraints are chosen based on the community's experience with what kinds
of queries users wish to make on parse trees; in general, Tregex has more
emphasis on longer-scale ancestry and predecessor/successor relationships,
and less emphasis on fixed-arity nodes and the sequence of children,
compared to IDSgrep.  Tregex includes special features for manipulating
``heads,'' but they refer to the linguistic meaning of that term in relation
to parse trees, not the rather different EIDS-specific definition.

To make EIDS trees searchable with Tregex, it was necessary to translate the
trees into the syntax used by tree bank files, which expresses
variable-arity trees using nested parentheses and alphanumeric labels.  We
used a Perl script to do this translation, using identifiers starting with
A, B, C, and D for functors of nodes with heads (arity zero to three
respectively); U, V, W, and X for functors of nodes without heads; and H for
head values themselves, all those prefixes being followed by the hexadecimal
Unicode code point values, separating additional code points with
underscores in the case of multi-character strings.  Each node had a label
corresponding to its functor and arity, then the first child would be the
EIDS head if any, and any remaining children would be the ordered children
of the node in the EIDS tree.  We also inserted a special node with the
label R at the root, without which it might be difficult to avoid having all
queries function as match-anywhere.  Tregex was invoked using its command
line interface with the -o option, to have it return only one
result for each match, as opposed to its default of returning multiple
matches when a single tree can satisfy the query constraints in multiple
ways.

For example, the EIDS \eids{<明>⿰日月} was translated to the tree bank-style
tree (R (C2FF0 H660E (A3B H65E5) (A3B H6708))).  That can be read as a root
R with one child, which in turn is binary with functor U+2FF0 (\eids{⿰})
and head U+660E (\eids{明}).  The next two children in the tree bank-style
tree are both nullary with functor U+003B (semicolon, which is implicit in
the EIDS syntax), and heads U+65E5 (\eids{日}) and U+6708 (\eids{月})
respectively.

With this encoding scheme for the trees, almost all of our test query set
can be translated into Tregex queries by recursively expressing the match
condition at each node in Tregex terms.
\begin{itemize}
  \item Exact head-to-head matching is equivalent to checking the first
    grandchild of R for an exact match.  For instance, the EIDS query
    \eids{明} becomes ``R <, ( \_\_ <, H660E )''.
  \item Match-anywhere on a single character (assuming no headless
    semicolons occur in the
    dictionary, which is true of our test dictionaries) is equivalent to
    simply searching for the appropriate head token.  We include a check for
    the root node, to prevent Tregex from returning multiple matches within
    the same dictionary entry.  For instance, the EIDS
    query \eids{...明} becomes ``R <{}< H660E''.
  \item The headless dictionary entry test queries can be translated to
    Tregex queries by a simple recursion.  Some care is needed to match
    internal nodes both with and without heads.  For instance, the EIDS query
    \eids{⿰日月} becomes ``R <, ( C2FF0 <2 ( \_\_ <, H65E5 ) <3 ( \_\_ <,
    H6708 ) ) | <, ( W2FF0 <, ( \_\_ <, H65E5 ) <2 ( \_\_ <, H6708 ) )''.
  \item Dictionary entry queries with wildcards in the leaves can be formed
    by omitting the matching constraints for the wildcard nodes.
  \item Unordered match queries follow the same pattern as their ordered
    equivalents, substituting the general child operator ``<'' for
    position-specific child operators like ``<2''.
  \item Boolean OR, AND, and NOT, are directly supported in the Tregex query
    language.  For instance, the EIDS query \eids{\&...心!...日} becomes ``R
    <{}< H5FC3 !<{}<H65E5''.
\end{itemize}

Only the associative-match queries were omitted; queries of that general
nature could be done in Tregex with its predecessor and successor
operations, but to exactly match the semantics of IDSgrep's
associative-match operator would involve many additional constraints to
exclude exotic cases, such as matching nodes with the right functor but
wrong arity.  It is not clear how to perform a fair comparison between the
two systems on such queries: writing a query with exactly identical
semantics to IDSgrep would seem to penalize Tregex by making it do a great
deal of superfluous computation not necessary for correct results from
ordinary data, but tuning the queries to the data in a way that
significantly changed their semantics would render the comparison
meaningless except on the data for which the queries had been tuned.

We ran 20 loops of all test queries except the 53 associative-match queries
against the combined test dictionaries, using IDSgrep in its default
configuration, IDSgrep with match filtering turned off, and Tregex.  The
timing results (measured in seconds, with the means and sample standard
deviations over the 20 loops) are in Table~\ref{tab:tregex-comparison}.

\begin{table}
  \tbl{Timing comparison between IDSgrep and Tregex.}{
  \begin{tabular}{l|rr}
    search software & mean & std.\ dev. \\ \hline
    IDSgrep (default) & 151.36 & 0.81 \\
    IDSgrep (no filtering) & 991.50 & 0.23 \\
    Tregex & 5928.93 & 15.62
  \end{tabular}
  }
  \label{tab:tregex-comparison}
\end{table}


\subsection{Discussion}

Table~\ref{tab:overview} illustrates the differences between the four test
dictionaries.  On the 160 single-character searches, the CHISE and CJKVI-J
dictionaries each return fewer than 160 results, because these dictionaries
only contain entries for characters when they have nontrivial
decompositions.  The KanjiVG dictionary, however, derives from a data source
primarily concerned with the strokes rather than the component breakdown, so
it includes an entry for every character in its scope even if the breakdown
is trivial; and Tsukurimashou includes two entries (giving component
breakdown and source code information in separate entries) for every
character.

Bearing in mind that our test queries are derived from CJKVI-J entries,
the headless-entry and unordered-match queries return only a few results in
the other databases because of differences in how the dictionaries break
down the same characters.  That effect shows up more strongly with the
associative-match queries.  The 53 associative queries return 54 results
from CJKVI-J despite its canonicalization intended to make associative
matching unnecessary, because it contains separate entries for U+66F8 and
U+2F8CC, both of which look like \eids{書} and have the same decomposition. 
Even with associative matching, only a few of the queries in this class
return results from the other databases, because of differing breakdowns. 
Finally, note the similarity in all databases (nearly identical match counts)
between the headless-entry and unordered-match queries.  It appears to be a
property of the Han character set that there are very few pairs of
characters differing only by a reordering of subtrees at the root level (for
instance, swapping the left and right of a left-right character).

The timing and filter hit results in Table~\ref{tab:speed-test} show the
effect of filtering.  Lambda filtering (the simpler layer, which was
implemented first in IDSgrep's development) eliminates a little over half of
the tree tests given the distribution of queries and dictionary data we
used.  With tree tests accounting for most of the running time, lambda
filtering gives very close to a factor of two speed-up overall.  BDD
filtering eliminates 91.3\% of the tree tests and gives a factor of 5.75
speed-up.

However, there is little additional benefit to using both filters at once. 
Although we found the difference to be statistically significant, the sample
mean running time for both filters together is just 2.4\% faster than for
BDD filtering alone.  Note that the difference in raw number of BDD hits per
loop is only eight hits, on a total of almost 31 million; any tree match
avoided by the lambda filter would almost certainly be eliminated by the BDD
filter anyway, and the speed benefit from the lambda filter in this
configuration can be attributed to avoiding the BDD checks themselves.  Both
filter implementations exist in the current version of IDSgrep because of
the history of its development, but a new implementation might omit the
lambda filters without any important loss of speed.  On the other hand,
because it does not require an external BDD library, the lambda filter
implementation may still be useful in installations where the external
dependency is undesirable.

The correlation between the filters can be understood by considering how
they share their vector definitions.  We can imagine an ideal exact
filtering function of bit vectors that returns true exactly on those bit
vectors, and only those, which could possibly be associated with matching
trees.  Such a function would extract all possible information from the bit
vectors and give the best possible filtering given our vector-creating
function.  The lambda filtering function is a coarse one-sided approximation
of the ideal filtering function, but the BDD filter as implemented almost
perfectly approximates the ideal.  It gives false positives \emph{relative
to the ideal filtering function} only in the relatively rare cases where
implementation compromises in filter-calculus operations, like the
existential quantification of very complicated trees, force a loss of
precision.  To see a tree check eliminated by the lambda filter and not the
BDD filter, the tree check would have to not only be among the roughly 56\%
of non-hits that the lambda filter is able to eliminate at all, but also be
in the very small set of hits that differ between the implementation of BDD
filters and the hypothetical ideal bit-vector filter.  Conversely, to
achieve meaningfully better bit vector filtering in the sense of eliminating
more tree checks compared to the IDSgrep implementation of BDD filters, it
would be necessary to use better bit vectors (perhaps with more bits per
vector, for instance), not just a better filter on the same vectors.

Tree-match memoization is not expected to make much difference to practical
applications, but the experimental results on it illustrate the asymptotic
behaviour of the algorithm.  Applying increasing numbers of nested
match-anywhere operators slows down the matching linearly in the default
configuration (with memoization on demand); despite the worst-case bound of
$O(n^3)$ for the algorithm, the case tested in our experiment involves
checking a linearly-increasing query against a database that does not
change, with constant-time tests for each pair of nodes, so $\Theta(n)$
performance is what we might expect.  With memoization, the matching time
increases exponentially, also as we would expect from theory, although as
seen in Figure~\ref{fig:memo-times}, the fit there is less close.

Any comparison to other software is made more difficult by IDSgrep's unique
application domain, but GNU grep and Tregex are typical of what someone
without IDSgrep might use for similar purposes.  The CHISE
project~\cite{CHISE}, in particular, offers an online IDS search that is
essentially a substring search on its IDS database, as well as editor
plugins to support local regular expression searching similar to grep on the
same data.

For the simple match-anywhere and head-to-head single character queries we
tested, GNU grep is unquestionably much faster than IDSgrep, by a factor of
50.6 (in sample mean user CPU time per loop) compared to IDSgrep without
filtering, or 15.4 with filtering.  The comparison without filtering may be
more fair because GNU grep does not benefit from a precomputed index.  On
the other hand, IDSgrep is not really intended for this kind of query; its
goal is to answer detailed structural queries which GNU grep cannot do at
all.  It remains that IDSgrep might benefit from switching to a faster
string search algorithm from its current filtered tree match, when it can
detect that a query is of a simple form that could be answered by string
search.

For more advanced structural queries on trees, Tregex seems a reasonably
comparable package to IDSgrep.  Both are specialized to their application
domains, and they have different application domains, so they are not
perfectly comparable.  In our comparison, covering almost all of our
original query speed test set, IDSgrep was found to be 39.17 times as fast
as Tregex if allowed to use its precomputed bit vector indices, or 5.98
times as fast without them.  Factors contributing to the speed difference
may include the basic speed difference between IDSgrep's compiled C code and
Tregex's Java; differences between our test databases and the kind of data
Tregex more commonly uses (in particular, our encoding of EIDSes to tree
bank syntax resulted in many more unique tag values, and longer tag values,
than usually occur in parse trees); and the fact that Tregex solves a harder
problem.  Although our test queries did not use this feature directly, the
Tregex query language allows binding named variables to nodes in the tree
and applying Boolean constraints to them.  It is not difficult to show that
Tregex's matching problem is NP-hard by a reduction from 3SAT, in contrast
to IDSgrep's matching problem, which has a $O(n^3)$ time bound.

To summarize the comparison between programs, it would be reasonable to say
that GNU grep is designed for speed in preference to expressive power;
Tregex is designed for expressive power in preference to speed; and IDSgrep
falls somewhere in between.


\section{Conclusions and future work}
\label{sec:conclusion}

We have described the IDSgrep structural query system for Han character
dictionaries:  its data model, query language, and details of the algorithms
it uses, with experimental results.

IDSgrep was first developed to support Japanese-language font development in
the Tsukurimashou Project.  The user base in that application is small and
highly trained.  Accordingly, IDSgrep was designed to be powerful, and
convenient for an expert user, rather than easy for beginners to learn.  A
goal stated in the IDSgrep documentation was to achieve the
user-friendliness of conventional Unix grep.  But the concept of spatial
structure queries on Han characters is of potential interest to other users,
including many who are not computer scientists.  It is an open question to
what extent language learners and other dictionary users may find IDSgrep
useful; and, if the query language and command-line interface are in fact
obstacles, what could make spatial queries more accessible to non-experts
while retaining their power.  It is easy to imagine that some kind of
graphical query system could use IDSgrep internally, either by directly
calling the existing software or with a new implementation of the same or a
similar algorithm and data model.

The algorithmic ideas in IDSgrep may have more general application.  In
particular, BDD filtering of Bloom-style bit vectors is novel, at least to
the computational linguistics domain, and may be a useful extension to
existing bit vector techniques for parsing of unification-based grammars. 
Its application to this and other problems beyond character dictionary
search is another possible direction for future work.


\bibliographystyle{ACM-Reference-Format-Journals}
\bibliography{idsgrep}


\end{document}